%% file: main.tex
\documentclass[10pt,journal,compsoc]{IEEEtran}

\usepackage{soul}
\usepackage{url}
\usepackage{hyperref}
\usepackage{graphicx}
\usepackage{amsmath}
\usepackage{booktabs}

\usepackage{subfigure}
\usepackage{bm}
\usepackage{mathrsfs}
\usepackage{enumerate}
\usepackage{enumitem}
\usepackage{multirow}
\usepackage{amsfonts,amssymb}
\usepackage{tabularx}
\usepackage[ruled,linesnumbered]{algorithm2e}
\usepackage{algpseudocode}
\usepackage{color}

\ifCLASSOPTIONcompsoc
  \usepackage[nocompress]{cite}
\else
  \usepackage{cite}
\fi

\ifCLASSINFOpdf
\else
\fi

\begin{document}
%
\renewcommand{\algorithmicrequire}{\textbf{Input:}}  
\renewcommand{\algorithmicensure}{\textbf{Output:}} 

\title{Towards Robust Knowledge Graph Embedding via Multi-task Reinforcement Learning}

\author{Zhao Zhang, 
	Fuzhen Zhuang, 
	Hengshu Zhu,~\IEEEmembership{Senior Member,~IEEE,} \\
	Chao Li,
	Hui Xiong,~\IEEEmembership{Fellow,~IEEE,}
	Qing He,
	and Yongjun Xu
\IEEEcompsocitemizethanks{
	\IEEEcompsocthanksitem Zhao Zhang, Qing He and Yongjun Xu are  
	with Institute of Computing Technology, Chinese Academy of Sciences, Beijing 100190, China. Zhao Zhang is also with Zhejiang Lab, Hangzhou, China. \\
	Email: \{zhangzhao2021, heqing, xyj\}@ict.ac.cn
	\IEEEcompsocthanksitem Fuzhen Zhuang is with Institute of Artificial Intelligence, Beihang University, Beijing 100191, China, and SKLSDE, School of Computer Science, Beihang University, Beijing 100191, China. \\
	Email: zhuangfuzhen@buaa.edu.cn
	\IEEEcompsocthanksitem Hengshu Zhu is with Baidu Talent Intelligence Center, Beijing
	100085, China. \\
	Email: zhuhengshu@baidu.com
	\IEEEcompsocthanksitem Chao Li is with Zhejiang Lab, Hangzhou, China, and Institute of Computing Technology, Chinese Academy of Sciences, Beijing 100190, China. \\
	Email: lichao@zhejianglab.com
	\IEEEcompsocthanksitem Hui~Xiong is with Artificial Intelligence Thrust, The Hong Kong University of Science and Technology, Guangzhou 511458, China. \\
	E-mail: xionghui@ust.hk
	\IEEEcompsocthanksitem Fuzhen Zhuang and Chao Li are corresponding authors.
	}
}

\IEEEtitleabstractindextext{%
\begin{abstract}
Nowadays, Knowledge graphs (KGs) have been playing a pivotal role in AI-related applications. Despite the large sizes, existing KGs are far from complete and comprehensive.
In order to continuously enrich KGs, automatic knowledge construction and update mechanisms are usually utilized, which inevitably bring in plenty of noise. However, most existing knowledge graph embedding (KGE) methods assume that all the triple facts in KGs are correct, and project both entities and relations into a low-dimensional space without considering noise and knowledge conflicts. This will lead to low-quality and unreliable representations of KGs. To this end, in this paper, we propose a general multi-task reinforcement learning framework, which can greatly alleviate the noisy data problem. In our framework, we exploit reinforcement learning for choosing high-quality knowledge triples while filtering out the noisy ones. Also, in order to take full advantage of the correlations among semantically similar relations, the triple selection processes of similar relations are trained in a collective way with multi-task learning. Moreover, we extend popular KGE models TransE, DistMult, ConvE and RotatE with the proposed framework. Finally, the experimental validation shows that our approach is able to enhance existing KGE models and can provide more robust representations of KGs in noisy scenarios.
\end{abstract}

}

\maketitle

\IEEEdisplaynontitleabstractindextext

%
\IEEEpeerreviewmaketitle

\input{introduction}

\input{related_work}
\input{model}

\input{experiments}

\input{conclusion}

\bibliographystyle{IEEEtran}
\bibliography{TKDE_Noise}

\appendices

\section{The Impact of the Clustering Algorithm on the Final Results}

\begin{figure}[th]
	\center
	\includegraphics[width=8cm,height=6cm]{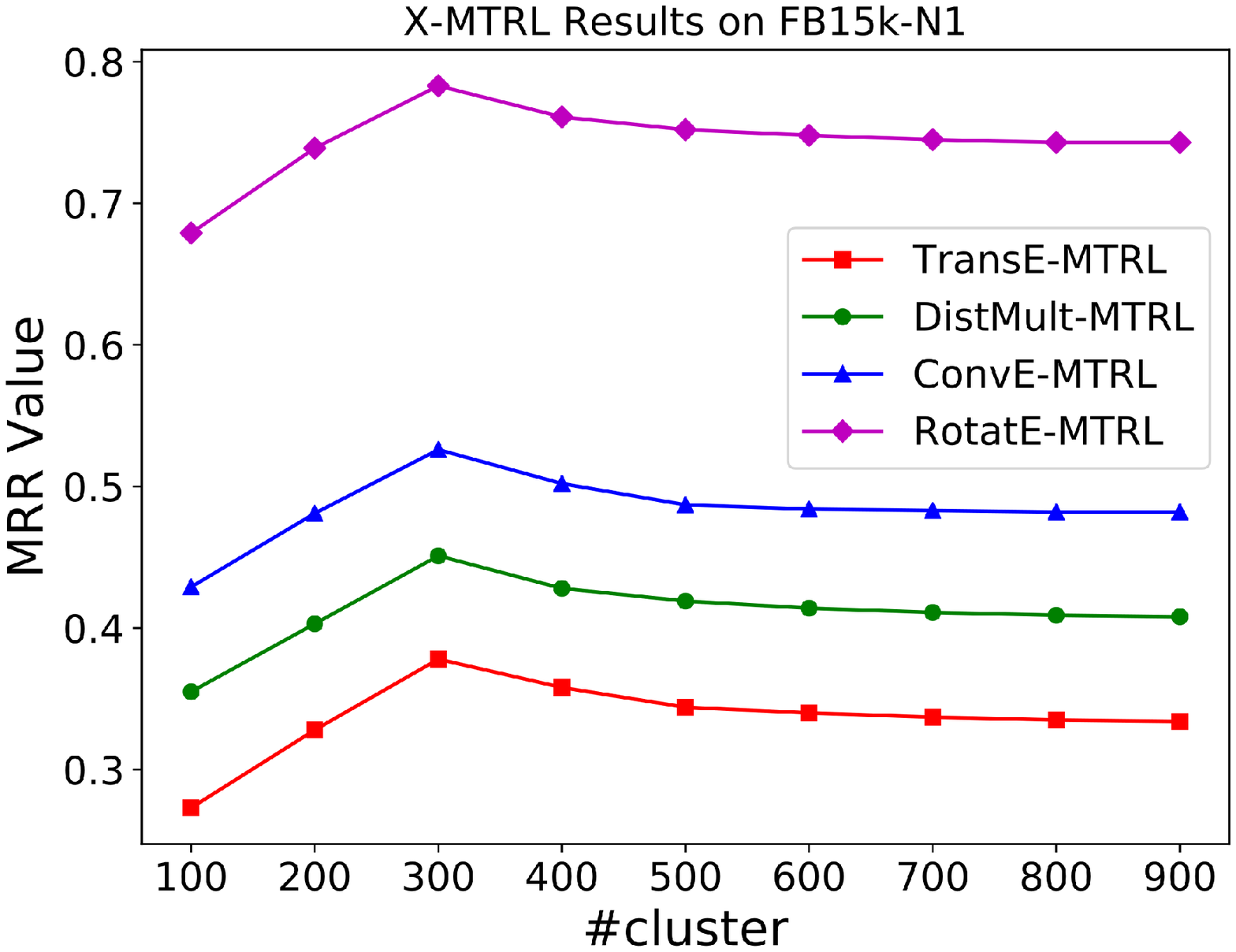}
	\caption{\small{The impact of the number of clusters on the results of X-MTRL on FB15k-N1.}}
	\label{fig:fb15k_cluster}
\end{figure}

\begin{figure}[th]
	\center
	\includegraphics[width=8cm,height=6cm]{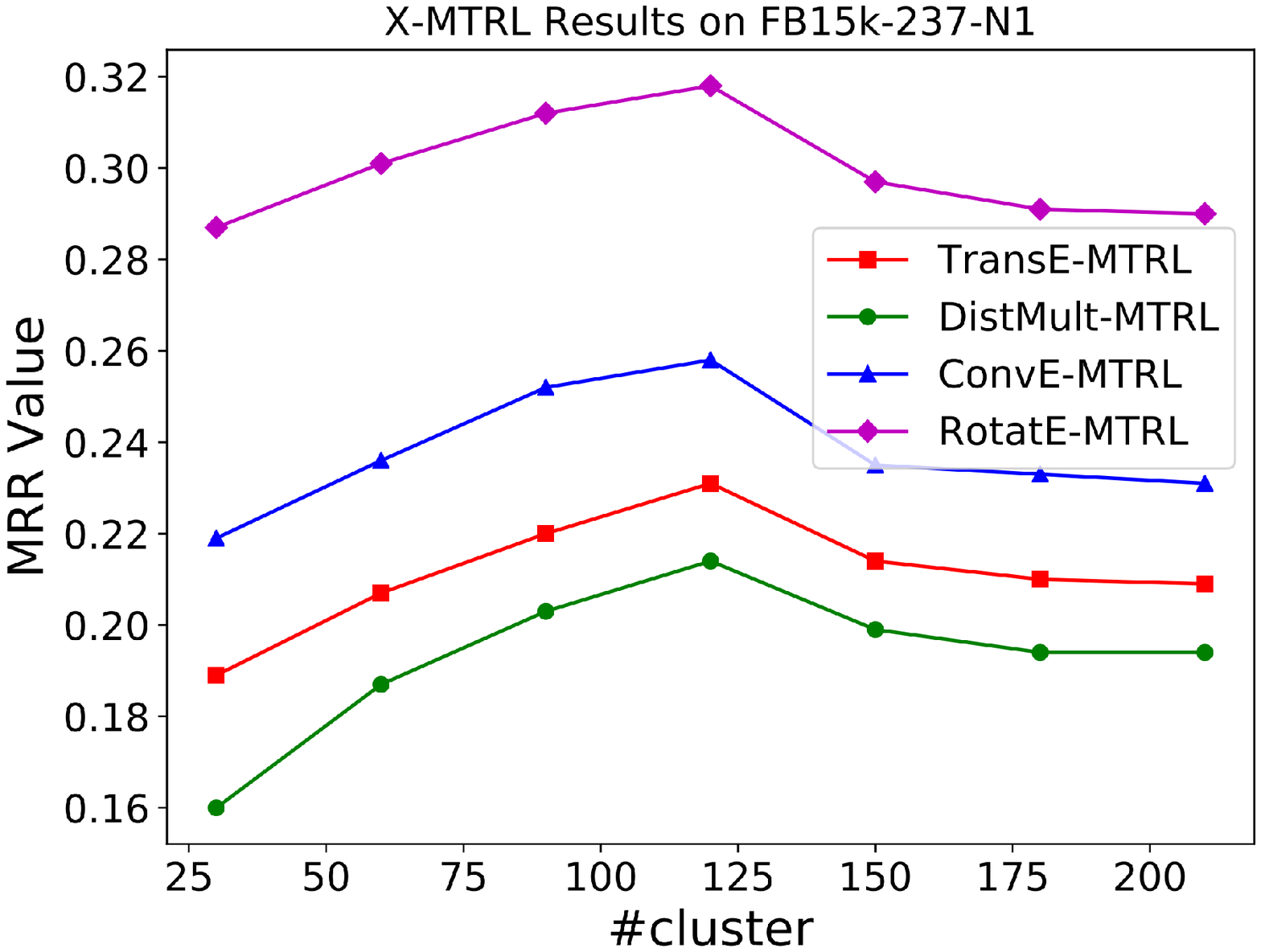}
	\caption{\small{The impact of the number of clusters on the results of X-MTRL on FB15k-237-N1.}}
	\label{fig:fb15k237_cluster}
\end{figure}

\begin{figure}[th]
	\center
	\includegraphics[width=8cm,height=6cm]{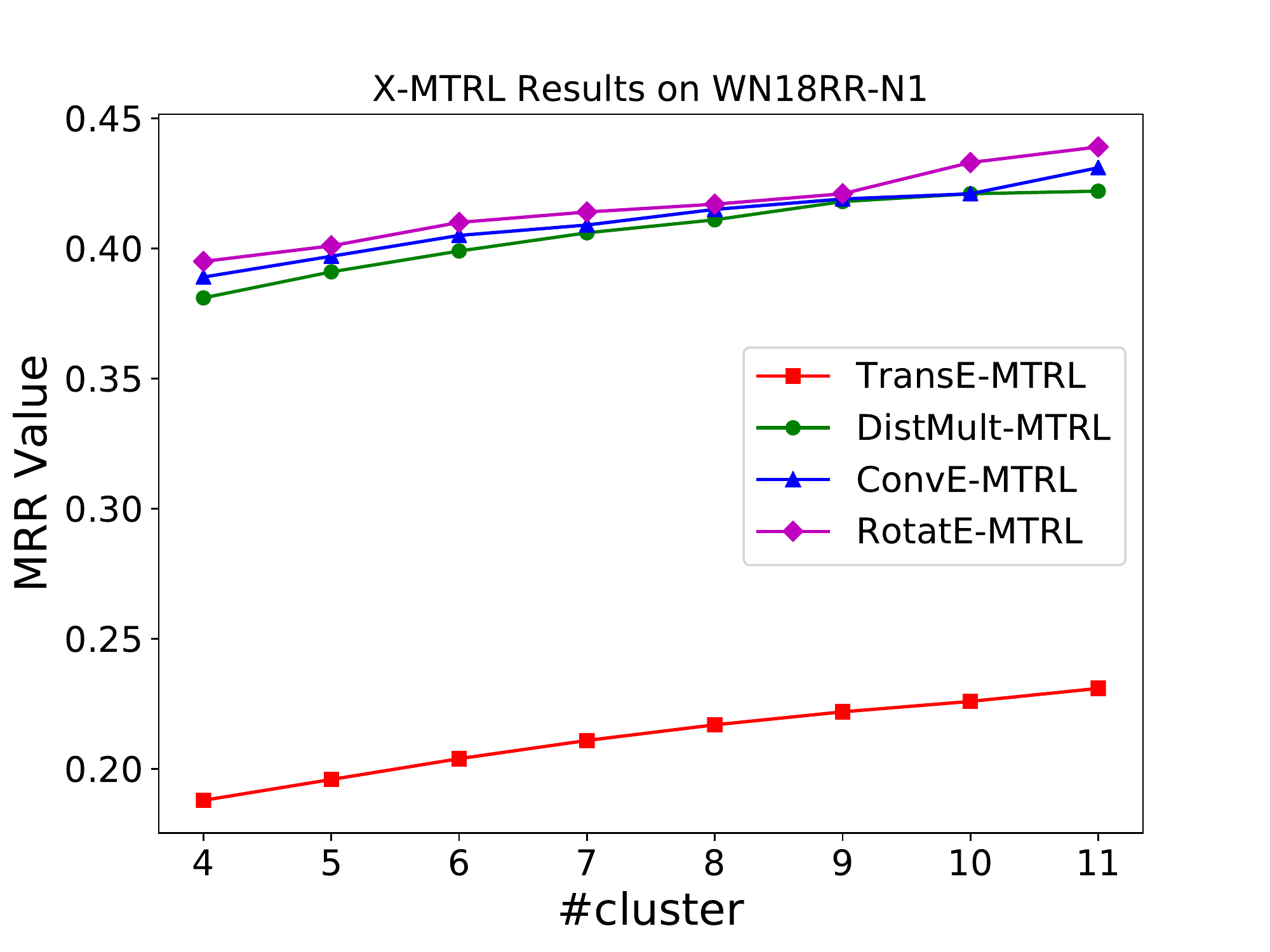}
	\caption{\small{The impact of the number of clusters on the results of X-MTRL on WN18RR.}}
	\label{fig:wn18rr_cluster}
\end{figure}

In this section, we analyze the impact of the clustering algorithm on the final results. In this paper, we use the k-means algorithm to cluster relations into different groups. Figure~\ref{fig:fb15k_cluster}, Figure~\ref{fig:fb15k237_cluster} and Figure~\ref{fig:wn18rr_cluster} show the impact of the relation cluster number on the final results. We have the following findings. 

(1) There exists an optimal value for the number of relation clusters on FB15k and FB15k-237 based datasets. The models keep achieving better results as the number of relations goes from 0 to the optimal value. Then, after the value exceeds the optimal point, the results start falling down to a stable value. The reason lies as: (i) a small value of relation clusters leads to large-sized relation clusters. In this case, some unrelated relations may join in the same cluster, and degrade the performance; (ii) a large value of relation clusters leads to small-sized relation clusters, thus each relation cannot take full advantage of the information from semantically related relations. In this case, the results are also unsatisfactory.

(2) On WN18RR-based datasets, the results keep going up as the number of relation clusters increases. The reason is that the 11 relations in WN18RR are semantically unrelated. The information from semantically related relations is able to benefit the results, and the information from unrelated relations may degrade the performance. 
Our findings again validate that the MTRL model is more useful for
KGs which have dense semantic distributions over relations,
while the STRL model is more suitable for KGs in which the
semantic correlations among relations are weak.
It is worth noting that we also tried different random seeds to run the clustering algorithm, but find the seed does not have a big effect on the final results (up to 0.001 MRR score).

\section{Link Prediction Results on State-of-the-art KGE models}
To further evaluate the generalization performance of the proposed framework, we extend popular knowledge graph embedding (KGE) models R-GCN~\cite{schlichtkrull2018modeling}, CompGCN~\cite{vashishth2020composition} and ComplEx-N3~\cite{lacroix2018canonical} with our framework.
The results are shown in Table~\ref{tab:kgc_results_fb15k_other} to Table~\ref{tab:kgc_results_wn18rr_other}.
Results in bold font are the best results in the group, and the underlined results denote the best results in the column. 
Numbers marked with * indicate that the improvement is statistically significant compared with the best baseline in the group (t-test with p-value \textless 0.05).
Experimental results show that our extended models again outperform baseline competitors, which validate the generalization ability of the proposed framework.

\begin{table*}[t]
	\centering
	\caption{Link prediction results on FB15k-N1 to FB15k-N3.}
	\label{tab:kgc_results_fb15k_other}
	\scalebox{1}{
		\begin{tabular}{@{}lcccccccccccc@{}}
			\toprule
			& \multicolumn{4}{c}{FB15k-N1} & \multicolumn{4}{c}{FB15k-N2} & \multicolumn{4}{c}{FB15k-N3}
			\\
			\cmidrule(lr){2-5} \cmidrule(l){6-9} \cmidrule(l){10-13} & MRR & H@10 & H@3 & H@1 & MRR & H@10 & H@3 & H@1 & MRR & H@10 & H@3 & H@1  \\
			\midrule
			
			CKRL & 0.371 & 0.642 & 0.440 & 0.229 & 0.349 & 0.611 & 0.421 & 0.211 & 0.317 & 0.566 & 0.378 & 0.190 \\
			\midrule
			
			R-GCN & 0.654 & 0.786 & 0.627 & 0.532 & 0.603 & 0.732 & 0.584 & 0.498 & 0.531 & 0.701 & 0.535 & 0.429 \\
			R-GCN-NoiGAN & 0.661 & 0.788 & 0.634 & 0.543 & 0.605 & 0.741 & 0.592 & 0.501 & 0.534 & \textbf{0.722} & 0.551 & 0.439 \\
			R-GCN-STRL (Ours) & 0.669 & 0.788 & \textbf{0.642} & 0.544 & 0.611 & 0.739 & 0.599 & 0.512 & 0.535 & 0.712 & 0.556 & 0.441 \\
			R-GCN-MTRL (Ours) & \textbf{0.674*} & \textbf{0.795*} & 0.639 & \textbf{0.552*} & \textbf{0.619*} & \textbf{0.753*} & \textbf{0.603*} & \textbf{0.521*} & \textbf{0.539*} & 0.719 & \textbf{0.568*} & \textbf{0.453*} \\
			\midrule
			
			CompGCN & 0.792 & 0.845 & 0.827 & 0.744 & 0.761 & 0.826 & 0.793 & 0.713 & 0.743 & 0.792 & 0.771 & 0.682 \\
			CompGCN-NoiGAN & 0.793 & \textbf{0.849} & 0.832 & 0.743 & 0.759 & 0.821 & 0.787 & 0.718 & 0.746 & 0.791 & 0.775 & 0.689 \\
			CompGCN-STRL (Ours) & 0.796 & 0.846 & 0.838 & 0.746 & 0.764 & 0.831 & 0.799 & 0.717 & 0.749 & \textbf{0.804} & 0.781 & 0.693 \\
			CompGCN-MTRL (Ours) & \textbf{0.801*} & 0.847 & \textbf{0.842*} & \textbf{0.749*} & \textbf{0.771*} & \textbf{0.833*} & \textbf{0.806*} & \textbf{0.722*} & \textbf{0.762*} & 0.801 & \textbf{0.798*} & \textbf{0.701*} \\
			\midrule
			
			ComplEx-N3 & 0.795 & 0.844 & 0.822 & 0.753 & 0.754 & 0.813 & 0.783 & 0.702 & 0.745 & 0.795 & 0.768 & 0.691 \\
			ComplEx-N3-NoiGAN & 0.799 & 0.853 & 0.829 & 0.755 & 0.757 & 0.822 & 0.779 & 0.713 & 0.751 & 0.793 & 0.771 & 0.696 \\
			ComplEx-N3-STRL (Ours) & 0.804 & 0.855 & 0.828 & 0.759 & 0.763 & 0.827 & 0.783 & 0.712 & 0.755 & \textbf{0.803} & 0.777 & 0.704 \\
			ComplEx-N3-MTRL (Ours) & \textbf{0.807*} & \textbf{0.857*} & \textbf{0.839*} & \textbf{0.764*} & \textbf{0.771*} & \textbf{0.831*} & \textbf{0.794*} & \textbf{0.725*} & \textbf{0.765*} & 0.802 & \textbf{0.785*} & \textbf{0.709*} \\
			\bottomrule
			
		\end{tabular}
	}
\end{table*}

\begin{table*}[t]
	\centering
	\caption{Link prediction results on FB15k-237-N1 to FB15k-237-N3.}
	\label{tab:kgc_results_fb15k237_other}
	\scalebox{1}{
		\begin{tabular}{@{}lcccccccccccc@{}}
			\toprule
			& \multicolumn{4}{c}{FB15k-237-N1} & \multicolumn{4}{c}{FB15k-237-N2} & \multicolumn{4}{c}{ FB15k-237-N3}
			\\
			\cmidrule(lr){2-5} \cmidrule(l){6-9} \cmidrule(l){10-13} & MRR & H@10 & H@3 & H@1 & MRR & H@10 & H@3 & H@1 & MRR & H@10 & H@3 & H@1  \\
			\midrule
			
			CKRL & 0.227 & 0.387 & 0.249 & 0.144 & 0.209 & 0.371 & 0.234 & 0.133 & 0.195 & 0.359 & 0.222  & 0.129 \\
			\midrule
			
			R-GCN & 0.203 & 0.358 & 0.233 & 0.127 & 0.192 & 0.344 & 0.221 & 0.114 & 0.177 & 0.335 & 0.202  & 0.102 \\
			R-GCN-NoiGAN & 0.207 & 0.361 & 0.236 & 0.122 & 0.193 & 0.349 & 0.224 & 0.115 & 0.178 & 0.339  & \textbf{0.209} & 0.109 \\
			R-GCN-STRL (Ours) & 0.211 & 0.366 & \textbf{0.246} & 0.131 & 0.198 & 0.347 & 0.229 & 0.118 & 0.182 & 0.342 & 0.203 & 0.108 \\
			R-GCN-MTRL (Ours) & \textbf{0.215*} & \textbf{0.372*} & 0.241 & \textbf{0.139*} & \textbf{0.202*} & \textbf{0.355*} & \textbf{0.237*} & \textbf{0.126*} & \textbf{0.189*} &  \textbf{0.351*} & 0.207 & \textbf{0.122*} \\
			\midrule
			
			CompGCN & 0.308 & 0.497 & 0.347 & 0.199 & 0.304 & 0.479 & 0.317 & 0.181 & 0.277 & 0.472 & 0.309 & 0.176 \\
			CompGCN-NoiGAN & 0.311 & 0.495 & \textbf{0.349} & 0.203 & 0.311 & 0.485 & 0.327 & 0.198 & 0.287 & 0.482  & 0.319 & 0.181 \\
			CompGCN-STRL (Ours) & 0.313 & 0.503 & 0.341 & 0.212 & 0.309 & 0.484 & 0.331 & 0.194 & 0.283  & 0.486 & 0.322 & 0.193 \\
			CompGCN-MTRL (Ours) & \textbf{0.319*} & \textbf{0.508*} & 0.346 & \textbf{0.221*} & \textbf{0.315*} & \textbf{0.497*} & \textbf{0.342*} & \textbf{0.204*} & \textbf{0.309*}  & \textbf{0.497*} & \textbf{0.335*} & \textbf{0.201*} \\
			\midrule
			
			ComplEx-N3 & 0.311 & 0.505 & 0.341 & 0.193 & 0.309 & 0.482 & 0.322 & 0.188 & 0.281 & 0.477  & 0.316 & 0.182 \\
			ComplEx-N3-NoiGAN & 0.313 & 0.505 & 0.347 & 0.193 & 0.311 & 0.488 & 0.323 & 0.191 & 0.283 & 0.472 & 0.321 & 0.191 \\
			ComplEx-N3-STRL (Ours) & 0.322 & 0.509 & 0.354 & 0.204 & 0.313 & 0.491 & 0.331 & \textbf{0.203} & 0.292  & 0.486 & 0.329 & 0.195 \\
			ComplEx-N3-MTRL (Ours) & \textbf{0.329*} & \textbf{0.519*} & \textbf{0.359*} & \textbf{0.222*} & \textbf{0.318*} & \textbf{0.498*} & \textbf{0.341*} & 0.202 & \textbf{0.311*}  & \textbf{0.489*} & \textbf{0.336*} & \textbf{0.201*} \\
			\bottomrule
			
		\end{tabular}
	}
\end{table*}

\begin{table*}[t]
	\centering
	\caption{Link prediction results on WN18RR-N1 to WN18RR-N3.}
	\label{tab:kgc_results_wn18rr_other}
	\scalebox{1}{
		\begin{tabular}{@{}lcccccccccccc@{}}
			\toprule
			& \multicolumn{4}{c}{WN18RR-N1} & \multicolumn{4}{c}{WN18RR-N2} & \multicolumn{4}{c}{WN18RR-N3}
			\\
			\cmidrule(lr){2-5} \cmidrule(l){6-9} \cmidrule(l){10-13} & MRR & H@10 & H@3 & H@1 & MRR & H@10 & H@3 & H@1 & MRR & H@10 & H@3 & H@1  \\
			\midrule
			
			CKRL & 0.221 & 0.496 & 0.443 & 0.141 & 0.213 & 0.479 & 0.422 & 0.114 & 0.189 & 0.458 & 0.392 & 0.101 \\
			\midrule
			
			R-GCN & 0.414 & 0.479 & 0.432 & 0.379 & 0.391 & 0.457 & 0.417 & 0.356 & 0.372 & 0.442 & 0.396 & 0.331 \\
			R-GCN-NoiGAN & 0.420 & 0.483 & 0.435 & 0.382 & 0.394 & 0.465 & 0.419 & 0.362 & 0.377 & 0.439  & 0.399 & 0.342 \\
			R-GCN-STRL (Ours) & \textbf{0.431*} & \textbf{0.494*} & \textbf{0.443*} & 0.384 & \textbf{0.419*} & \textbf{0.479*} & \textbf{0.428*} & \textbf{0.374*} & \textbf{0.391*} & \textbf{0.458*} & 0.406 & \textbf{0.359*} \\
			R-GCN-MTRL (Ours) & 0.424 & 0.488 & 0.439 & \textbf{0.389} & 0.408 & 0.475 & 0.423 & 0.366 & 0.383 & 0.447 & \textbf{0.412} & 0.352 \\
			\midrule
			
			CompGCN & 0.434 & 0.537 & 0.461 & 0.404 & 0.421 & 0.521 & 0.444 & 0.383 & 0.397 & 0.503 & 0.419 & 0.362 \\
			CompGCN-NoiGAN & 0.438 & 0.538 & 0.465 & 0.409 & 0.424 & 0.522 & 0.446 & 0.389 & 0.399 & 0.512 & 0.421 & 0.367 \\
			CompGCN-STRL (Ours) & \textbf{0.447*} & \textbf{0.541*} & \textbf{0.471*} & \textbf{0.421*} & \textbf{0.433*} & \textbf{0.537*} & \textbf{0.462*} & \textbf{0.399*} & \textbf{0.416*} & \textbf{0.528*} & \textbf{0.429*} & \textbf{0.377*} \\
			CompGCN-MTRL (Ours) & 0.442 & 0.540 & 0.466 & 0.411 & 0.429 & 0.529 & 0.451 & 0.395 & 0.404 & 0.519 & 0.439 & 0.371 \\
			\midrule
			
			ComplEx-N3 & 0.438 & 0.538 & 0.456 & 0.408 & 0.419 & 0.522 & 0.441 & 0.381 & 0.402 & 0.501 & 0.423 & 0.359 \\
			ComplEx-N3-NoiGAN & 0.439 & 0.542 & 0.455 & 0.412 & 0.422 & 0.527 & 0.441 & 0.388 & 0.406 & 0.505 & 0.429 & 0.364 \\
			ComplEx-N3-STRL (Ours) & \textbf{0.452*} & 0.544 & \textbf{0.473*} & \textbf{0.427*} & \textbf{0.432*} & \textbf{0.539*} & 0.449 & \textbf{0.403*} & \textbf{0.419*} & \textbf{0.521*} & \textbf{0.439*} & \textbf{0.377*} \\
			ComplEx-N3-MTRL (Ours) & 0.445 & \textbf{0.549} & 0.465 & 0.419 & 0.430 & 0.533 & \textbf{0.452} & 0.397 & 0.412 & 0.511 & 0.433 & 0.373 \\
			\bottomrule
			
		\end{tabular}
	}
\end{table*}

%




\end{document}

%% file: introduction.tex
\IEEEraisesectionheading{\section{Introduction}\label{sec:introduction}}
\IEEEPARstart{K}{nowledge} graphs (KGs) are multi-relational directed graphs composed of entities as nodes and relations as different types of edges. They represent information about real-world facts in the form of knowledge triples, which are denoted as ($h$, $r$, $t$), where $h$ and $t$ correspond to the head and tail entities and $r$ denotes the relation between them, e.g., (\textit{Donald Trump}, \textit{nationality}, \textit{USA}).

Due to their effectiveness for representing structured data, knowledge graphs have been playing a pivotal role in various AI-related applications, including information retrieval~\cite{dalton2014entity}, question answering~\cite{ferrucci2010building}, and information extraction~\cite{mintz2009distant}. However, the underlying symbolic nature of knowledge triples often makes KGs hard to manipulate. Therefore, the recent attention has been drawn on knowledge graph embedding (KGE), which aims to project both entities and relations into a continuous low-dimensional space, so as to simplify the manipulation while preserving the inherent structure of the KGs. Such embeddings encode rich information of KGs, and are widely utilized by downstream applications~\cite{weston2013connecting,bordes2013translating,lin2015learning,wang2018ripplenet}.

\begin{figure}[t]
	\center
	\includegraphics[width=9cm,height=0.9cm]{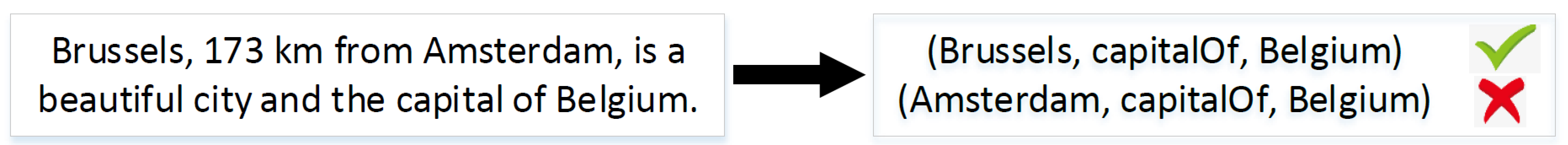}
	\caption{An example of relation extraction results.}
	\label{fig:RE_example}
\end{figure}

\begin{figure*}[t]
	\center
	\includegraphics[width=14cm,height=2.8cm]{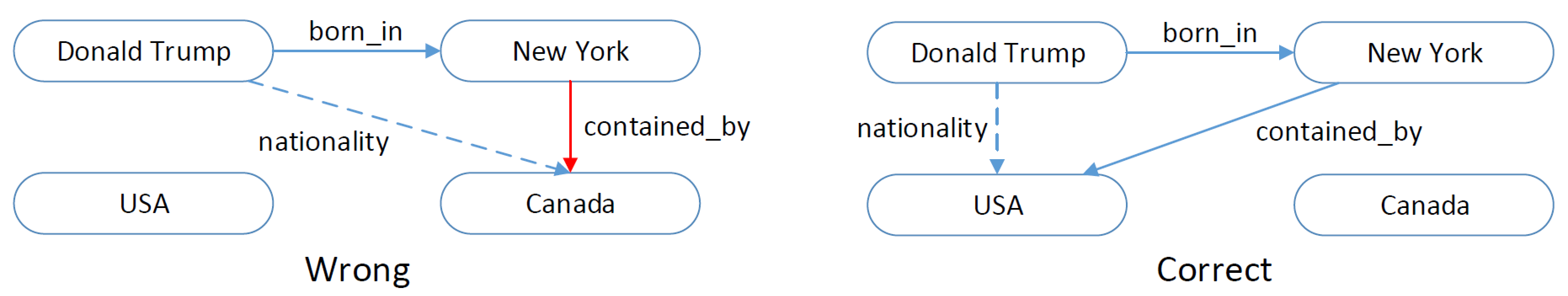}
	\caption{A knowledge graph completion case.}
	\label{fig:noise_case}
\end{figure*}

Even though the sizes of modern KGs are quickly growing, existing KGs, such as Freebase~\cite{bollacker2008freebase}, Wordnet~\cite{miller1995wordnet}, Yago~\cite{suchanek2007yago}, NELL~\cite{carlson2010toward}, Google's KG~\footnote{https://developers.google.com/knowledge-graph} and other domain-specific KGs such as Gene Ontology~\cite{ashburner2000gene}, are far from complete and comprehensive.
In order to continuously enrich KGs with the innumerable world knowledge, automatic mechanisms are utilized, which inevitably bring in plenty of noise and conflicts. 
Indeed, existing relation extraction models are not perfect~\cite{bai2019structured}. \cite{xie2018does} shows that state-of-the-art relation extraction models only achieve around 60\% precision when the recall is 20\%. 
Figure~\ref{fig:RE_example} shows an example of the relation extraction results.
When aiming to obtain the positive triple (\textit{Brussels}, \textit{capitalOf}, \textit{Belgium}), the negative triple (\textit{Amsterdam}, \textit{capitalOf}, \textit{Belgium}) is extracted, which indicates that the automatic mechanisms may bring in noise when enriching KGs.
Moreover, a novel task named Wikidata vandalism~\cite{heindorf2016vandalism}, which revealed the deliberate destructions that exist in KGs, has attracted wide attention. 

Indeed, the noisy data problem would degrade the performance of KGE models, thus lead to dissatisfactory results in downstream applications. Figure~\ref{fig:noise_case} shows a KG completion example, where KG completion aims to obtain new knowledge triples based on the existing ones in KGs with entity and relation embeddings. The solid lines represent existing triples, while the dashed lines denote the predicted ones. The left part of Figure~\ref{fig:noise_case} contains a noise triple (\textit{New York}, \textit{contained\_by}, \textit{Canada}) (denoted in a red line), and results in an incorrect prediction. The right part shows the correct prediction. Figure~\ref{fig:noise_case} shows that noisy triples in KGs greatly influence the performance of KGE models. 
However, this noisy data problem has largely been ignored by most of the existing models. 


Two recent studies focused on computing a confidence score for each triple~\cite{xie2018does}\cite{anonymous2020noigan}, and triples with higher confidence scores play a more important role during the training procedure. 
Particularly, CKRL~\cite{xie2018does} is the first work that aims to detect possible noise in KGs while learning knowledge representations with confidence simultaneously, which learns the confidence of a triple with both triple information and path information in KGs. Results show that by assigning different confidence scores to triples, KGE models obtain better performance.
NoiGAN~\cite{anonymous2020noigan} is another state-of-the-art noise-aware KGE model that unifies the task of noise detection and knowledge representation with a generative adversarial networks (GAN) framework. NoiGAN finds that assigning hard confidence scores, i.e., 0 or 1, to triples instead of soft ones is beneficial to achieve better results. 
In this paper, we also agree with NoiGAN that the correctness of triples should be treated with a hard decision, i.e., true or false, instead of being dealt with a soft confidence score. Since from an optimal view, positive triples should be fully leveraged and negative triples should be completely removed.



To this end, in this paper, we propose a novel multi-task reinforcement learning framework for robust KGE. Specifically, we first design reinforcement learning agents to select golden triples from the noisy training set while removing the false positives. Then a KGE model is trained based on the cleansed training set, and provides a delayed reward based on the quality of the selected triples for the reinforcement learning agents. Furthermore, it has been shown that semantically similar relations exist in large-scale KGs~\cite{zhang2018knowledge}. In order to take full advantage of the correlations among similar relations, we treat the triple selection process of each relation as a single task, and train the selection processes of semantically similar relations in a collective way with multi-task learning.
The reinforcement learning agents and the KGE model are trained in a joint way instead of independently, which avoids the accumulation of errors during the training process.


Our general framework can be easily utilized to extend a number of KGE models. Particularly, in this paper, we extend popular KGE models TransE~\cite{bordes2013translating}, DistMult~\cite{yang2015embedding}, ConvE~\cite{dettmers2018convolutional} and RotatE~\cite{sun2018rotate} with our framework. Extensive experiments on popular benchmarks demonstrate the effectiveness of our framework.

In a nutshell, we highlight our key contributions as follows,
\begin{enumerate}[leftmargin=*]
	\item We propose a general multi-task reinforcement learning framework for robust KGE. 
	\item Our proposed framework is able to extend a number of state-of-the-art KGE models without additional information like text or logical rules. Particularly, we extend popular KGE models TransE, DistMult, ConvE and RotatE in this paper.
	\item We evaluate our models on noisy datasets, and experimental results show that our extended models substantially outperform the base models as well as other baseline competitors.
\end{enumerate}

%% file: related_work.tex
\section{Related Work}\label{sec:related_work}

\subsection{Knowledge Graph Embedding}
Recent years have witnessed the increasing interest in KGE, which aims to represent entities and relations in KGs as low-dimensional vectors. Prior work roughly falls into three categories.
\begin{itemize}
\item Translation or rotation based models, which view relations as translations or rotations from a head entity to a tail entity~\cite{bordes2013translating,wang2014knowledge}. TransE~\cite{bordes2013translating} is one of the most widely used KGE model, which assumes $\mathbf{h} + \mathbf{r} \approx \mathbf{t}$ when $(h, r, t)$ holds. TransH~\cite{wang2014knowledge} is an extension of TransE, and introduces a mechanism of projecting entities into relation-specific hyperplanes that enables different roles of an entity in different relations. TransR~\cite{lin2015learning} introduces relation-specific projection matrices, and enables each entity to play different roles when involved in triples with different relations. STransE~\cite{nguyen2016stranse} further extends TransR and uses two projection matrices for each relation, which distinguish the role of each entity when acting as the head or tail of a triple. RotatE~\cite{sun2018rotate} projects entities and relations into a complex space, and view relations as rotations from head entities to tail entities. 
Rotate3D~\cite{gao2020rotate3d} models the
non-commutative composition pattern in three-dimensional
space with quaternion representation

\item Tensor factorization based models, which assume the score of a ($h$, $r$, $t$) can be factorized into several tensors~\cite{yang2015embedding,trouillon2016complex}. RESCAL~\cite{nickel2011three} is a representative model in the category, which represents each relation as a square matrix. The score function of RESCAL is defined as $f(h, r, t)=\mathbf{h}^\top \mathbf{M}_r \mathbf{t}$, where $f(h, r, t)$ denotes the score of triple $(h, r, t)$, $\mathbf{M}_r$ is a relation-specific matrix. Along this line, DistMult~\cite{yang2015embedding} simplifies RESCAL by restricting $\mathbf{M}_r$ as a diagonal matrix. ComplEx~\cite{trouillon2016complex} further extends DistMult, and projects both entities and relations into complex vectors instead of real-valued ones. 

\item Neural network based models, which leverage the power of deep neural networks or graph neural networks in representation learning to embed KGs. ConvE~\cite{dettmers2018convolutional} for the first time utilizes convolutional neural network (CNN) to capture the interactions between entities and relations. ConvKB~\cite{nguyen2018novel} models the relationships among the same dimensional entries of the embeddings. InteractE~\cite{vashishth2020interacte} extends ConvE by adding more interactions between entities and relations. R-GCN~\cite{schlichtkrull2018modeling} and KBAT~\cite{nathani2019learning} are two state-of-the-art models that adopt graph neural network to model relational data, where R-GCN uses graph convolutional network, and KBAT utilizes graph attention network.
\end{itemize}

In addition, besides the triple information in KGs, some studies also use external information like text~\cite{xie2016representation,Xu2017Knowledge} or logical rules~\cite{guo2016jointly,guo2018knowledge} to conduct the KGE task. 
It is worth noting that the four models, TransE, DistMult, ConvE and RotatE, which we extend in this paper with our framework, have covered the above three categories.
%
 
\subsection{Knowledge Graph Noise Detection}
Noise data inevitably exists in existing KGs. Most KG noise detection works rely on a large amount of human supervision. 
DBpedia~\cite{auer2007dbpedia} employs a worldwide crowd-sourcing effort to map knowledge triples to Wikipedia~\footnote{https://www.wikipedia.org/} info boxes. YAGO2~\cite{hoffart2013yago2} and Wikidata~\cite{vrandevcic2014wikidata} rely on human supervision to approve or reject a statement. These methods require a great deal of human supervision, which is extremely labor-intensive, time-consuming and, most importantly, usually unavailable in real-world scenarios. 
UKGE~\cite{chen2019embedding} aims to predict the confidence score for each triple. However, the setting of UKGE is different from our work. Specifically, in UKGE, each triple has a ground truth confidence score, while our work do not have such information.
Recently, CKRL~\cite{xie2018does} and NoiGAN~\cite{anonymous2020noigan} are proposed to detect noise in KGs with only internal information, which focused on computing a confidence score for each triple, and triples with higher confidence scores play a more important role during the training process of KGE models. CKRL~\cite{xie2018does} is the first noise-aware KGE model, which learns the confidence of a triple with both triple information and path information in KGs. NoiGAN~\cite{anonymous2020noigan} proposes a GAN-based framework that unifies the task of KG noise detection and knowledge representation learning. In this paper, we filter out the false positives with a hard decision, i.e., true or false, via multi-task reinforcement learning, and train the KGE models with the cleansed training data. 

\subsection{Multi-task Learning and Reinforcement Learning in KG-related Applications }
Previous studies have shown the benefits of applying multi-task learning to KG-related tasks.
Zhang et al.~\cite{zhang2018multie}\cite{zhang2018knowledge} found semantically similar entities and relations in KGs, and trained the embeddings of similar entities and relations in a collective way with multi-task learning. Wang et al.~\cite{wang2019multi} proposed to use multi-task learning to jointly learn the entity representations in KGs and item representations in recommender systems. 
Luan et al.~\cite{luan2018multi} introduced a multi-task setup of identifying and classifying entities, relations, and coreference clusters in scientific articles.
Thus, this paper tries to train the triple selection processes of semantically similar relations in a collective way.

There are also some works that utilize reinforcement learning in KG-related tasks. Xiong et al.~\cite{xiong2017deeppath} used reinforcement learning in the path-based KG reasoning task. Along this way, Lin et al.~\cite{lin2018multi} further proposed to use the reward shaping technique to improve the results. Reinforcement learning has also been shown to beneficial in the relation extraction task, Feng et al.~\cite{feng2018reinforcement} and Qin~\cite{qin2018robust} used different settings to clean the training set in the distant relation extraction tasks. In this paper, we propose to use reinforcement learning to select positive triples from noisy training datasets for KGE models. To the best of our knowledge, this is the first work that applies reinforcement learning to filter noise in the KGE task.


%% file: model.tex
\vspace{-2mm}
\section{Methodology}\label{sec:methodology}
\vspace{-1mm}
In this section, we introduce the technical details of our framework. 
\vspace{-3mm}
\subsection{Overview}
\begin{figure*}
	\center
	\includegraphics[width=11cm,height=7.5cm]{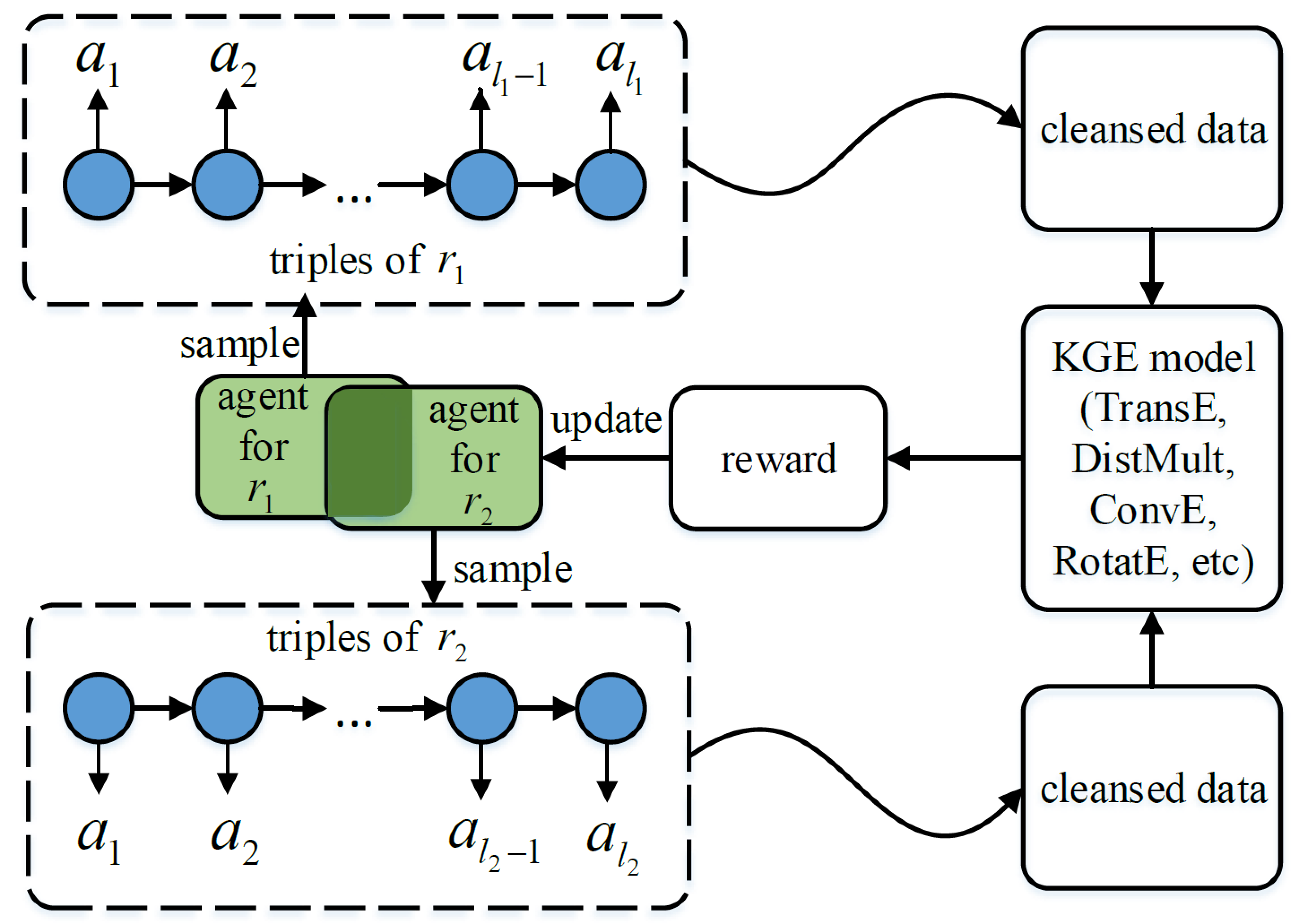}
	\caption{Reinforcement Learning Framework. $r_1$ and $r_2$ are semantically similar relations. $l_1$ and $l_2$ are the number of triples of $r_1$ and $r_2$ in the training set respectively.}
	\label{fig:rl_framework}
\end{figure*}

The proposed framework is shown in Figure \ref{fig:rl_framework}, where $r_1$ and $r_2$ are semantically similar relations. We define policy-based agents for $r_1$ and $r_2$ respectively. Since the two relations are semantically related, the two agents share some common information, which is denoted in dark green color in Figure \ref{fig:rl_framework}, making it possible for the two agents to be trained in a collective way. The reinforcement learning agents select golden triples for each relation, and provide a cleansed training set for the KGE model. At each state, the agent decides whether to select the current triple based on a stochastic policy. 
Then the KGE model is trained based on the cleansed training dataset, and provides a reward based on the model performance to update the policy-based agents.
In our framework, the KGE model can be substituted by a number of existing KGE models, which guarantees the flexibility and the extendibility  of our framework.
The policy-based agents and the KGE model are interleaved together and are trained in a joint way during the training process.
In the following, we introduce the details of the policy-based agents and the KGE model.

\subsection{Policy-based Agents}
We present the state, action, reward and the optimization details of the policy-based agents.
\subsubsection{State}
The state vector should encode the following information: (i) the relation corresponding to the current triple sequence; (ii) the triple when making decision on; (iii) the already selected triples. We represent the state at time step $t$ as a continuous real-valued vector $\mathbf{s}_t \in \mathbb{R}^{5d}$, which is calculated as  
\begin{equation}\label{eq:state} 
\mathbf{s}_t=\mathbf{r} \oplus \mathbf{h} \oplus \mathbf{t} \oplus \mathbf{\bar{h}} \oplus \mathbf{\bar{t}}  ,
\end{equation}
where $d$ is the embedding size of entities and relations, $\oplus$ is the concatenation operation for vectors. $\mathbf{r}$ is the embedding of the current relation. $\mathbf{h}$ and $\mathbf{t}$ are the embeddings of the head and tail entities from the current triple when making decision on. $\mathbf{\bar{h}}$ and $\mathbf{\bar{t}}$ are the average of embedding vectors for the head and tail entities from the already selected triples. All the parameters above are obtained from the KGE model. 

\subsubsection{Action}
The action space is $\{0,1\}$, where 1 indicates selecting the current triple, and 0 otherwise. Actions are decided based on a stochastic policy. We build an agent for each relation. Note that we also tried the setting that builds only one agent for all the relation types, but experiments show that we cannot get satisfactory performance under this setting. In this paper, we adopt the logistic function as the policy function:
\begin{align}
\pi_\Theta(a_t = 1|\mathbf{s}_t) &=\sigma(\mathbf{w}_r^\top \cdot \mathbf{s}_t) \label{eq:action1},\\
\pi_\Theta(a_t = 0|\mathbf{s}_t) &= 1 - \pi_\Theta(a_t = 1|\mathbf{s}_t) \label{eq:action0},
\end{align}
where $a_t$ denotes the action at time step $t$, $\sigma(\cdot)$ is the sigmoid function, $\mathbf{w}_r \in \mathbb{R}^{5d}$ is the policy parameter for relation $r$, $\Theta$ represents the parameters to be learned. It is worth noting that we also tried different network structures as our policy network including CNN and MLP, but find this simple setting achieves the best performance.

To take advantage of the correlations among semantically similar relations, inspired by the regularized multi-task learning algorithm~\cite{evgeniou2004regularized}, we decompose $\mathbf{w}_r$ into two parts, which is shown as 

\begin{equation}\label{eq:w_decompose} 
\mathbf{w}_r=\mathbf{u}_c + \mathbf{v}_r,
\end{equation}
where $\mathbf{u}_c \in \mathbb{R}^{5d}$ is the common model parameter for all the relations that belong to the same relation cluster, which facilitates the knowledge sharing among semantically similar relations. Meanwhile, $\mathbf{v}_r \in \mathbb{R}^{5d}$ is the specific parameter for each individual relation, which represents the different characteristics of each relation. 
In particular, we do not focus on how to obtain similar relations in this paper, thus we simply obtain relation clusters based on the results of TransE using the k-means algorithm, which is a simple relation clustering method introduced by \cite{zhang2018knowledge}.
The parameters of the policy function are denoted as $\Theta = \left \{ \mathbf{u}_1, \mathbf{u}_2, ..., \mathbf{u}_{|\mathcal{C}|} \right \} \cup  \left \{ \mathbf{v}_1, \mathbf{v}_2, ..., \mathbf{v}_{|\mathcal{R}|} \right \}$, where $\mathcal{C}$ and $\mathcal{R}$ are the relation cluster set and relation set respectively. $|\cdot|$ denotes the cardinality of a set.

\subsubsection{Reward}
We assume when the triple selection process for a relation is finished, the agent comes to a terminal state $\mathbf{s}_{|\mathcal{T}_r| + 1}$, where $\mathcal{T}_r$ is the triple set of relation $r$ in the training set. We get a delayed reward at the terminal state, while in other states, the reward is zero. We use the average score of the selected triples to compute the reward, which is
\begin{equation}\label{eq:reward} 
R=\frac{1}{|\hat{\mathcal{T}_r}|}\sum_{(h,r,t)\in \hat{\mathcal{T}_r}}f(h,r,t)+\alpha\frac{|\hat{\mathcal{T}_r}|}{|\mathcal{T}_r|},
\end{equation}
where $\hat{\mathcal{T}_r}$ is the set of selected triples for relation $r$. $f(h, r, t)$ is the utilized score function, which can be substituted by the score function of TransE, DistMult, ConvE or RotatE. $\alpha\frac{|\hat{\mathcal{T}_r}|}{|\mathcal{T}_r|}$ is a heuristic term to encourage the model to select more triples. 
Experiments show that if we don't add this term into the reward function, our model tends to select only a few high-score positive triples, while most positive triples are filtered out, making our framework unable to provide sufficient training data for the KGE model. Specifically, when $\hat{\mathcal{T}_r} = \varnothing$, the reward is set as the average score of all the training triples. This reward setting enables our model to effectively select golden triples from $\mathcal{T}_r$ and filter out $\mathcal{T}_r$ that only contains noisy triples.

\subsubsection{Optimization}
Our reinforcement learning model aims to maximize the expected reward for the triple selection process of each relation. More formally, for the triple selection process of relation $r$, the objective function is defined as 
\begin{equation}\label{eq:rl_loss} \vspace{0mm}
J(\Theta)=\mathbb{E}_{a\sim \pi_\Theta(a|\mathbf{s})}\left[R\right] + \Omega.
\end{equation}
The first term is the expected reward, and the second term is a regularized term inspired by the regularized multi-task learning algorithm~\cite{evgeniou2004regularized}:
\begin{equation}\label{eq:omega} 
\Omega=\lambda_1\left \| \mathbf{u}_c \right \|_2^2 + \lambda_2\left \| \mathbf{v}_r \right \|_2^2. 
\end{equation}
$\lambda_1$ and $\lambda_2$ are trade-off hyper-parameters. Large value of $\lambda_1$ will result in the separate training of each relation, while large value of $\lambda_2$ will lead to all relations in the same relation cluster sharing the same weight vector. Under this multi-task setting, the triple selection processes of relations that belong to the same relation cluster are trained in a collective way through sharing knowledge via $\mathbf{u}_c$.

According to the REINFORCE algorithm~\cite{williams1992simple}, we update the policy in the following way. For each relation, we sample a trajectory and get the corresponding reward $R$, and update the policy with the following gradients:
\begin{equation}\label{eq:gradients} 
\nabla_\Theta J(\Theta)=\sum_{t=1}^{|\mathcal{T}_r|}R \nabla_\Theta \log\pi_\Theta(a_t|\mathbf{s}_t) + \nabla_\Theta \Omega,
\end{equation}
where $\nabla$ represents the derivation operation.

In this paper, the extended KGE model using the proposed Multi-Task Reinforcement Learning framework is denoted as X-MTRL (X means TransE or DistMult or ConvE or RotatE in this paper).  Besides, we also provide a variant of the proposed framework by setting $\mathbf{u}_c$ in Equation (\ref{eq:w_decompose}) and Equation (\ref{eq:omega}) as $\mathbf{0}$, i.e., all the relations are trained separately and don't share any common information, which is a Single-Task Reinforcement Learning framework. The extended single-task model using the variant framework is denoted as X-STRL.

\subsection{KGE model}
Our framework is capable of extending a number of state-of-the-art KGE models, and in this paper, we apply our framework to four popular KGE models TransE, DistMult, ConvE, and RotatE.

TransE~\cite{bordes2013translating} is one of the most widely used KGE models, which views a relation as a translation from a head entity to a tail entity on the same low-dimensional hyperplane, i.e., $\mathbf{h} + \mathbf{r} \approx  \mathbf{t}$ when ($h,r,t$) holds. The score function of TransE is defined as \begin{equation}\label{eq:transe_score}
f(h,r,t)=-\left \| \mathbf{h}+\mathbf{r}-\mathbf{t} \right \|_{L_{n}}.
\end{equation}
$L_n$ can be $L_1$ or $L_2$ norm, which is decided based on the model performance over the validation set. Positive triples are supposed to have higher scores than negative ones. 

TransE adopts a margin-based loss function, which is defined as,
\begin{equation}\label{eq:margin_loss} 
\mathcal{L}_{TransE}=\sum_{(h,r,t) \in \hat{\mathcal{T}}}\left [ f(h^\prime,r,t^\prime)-f(h,r,t)+\gamma \right ]_+ ,
\end{equation}
where $\hat{\mathcal{T}}$ is the selected triples from the training set, $[\cdot]_+=\max(0, \cdot)$, and $\gamma$ is the margin separating positive instances from negative ones. $(h^\prime,r,t^\prime)$ represents negative triples which are generated by replacing the head entity or the tail entity of a positive triple
with a random entity in the KG. Formally, $(h^\prime,r,t^\prime) \in \left \{ (h^\prime,r,t)|h^\prime \in \mathcal{E} \right \}\cup \left \{ (h,r,t^\prime)|t^\prime \in \mathcal{E} \right \}$, where $\mathcal{E}$ is the entity set. 

DistMult~\cite{yang2015embedding} is a representative model in the tensor factorization category, which adopts a bilinear function to compute the scores of knowledge triples. The score function is defined as 
\begin{equation}\label{eq:distmult_score}
f(h,r, t) = \mathbf{h}^\top\mathbf{M}_r\mathbf{t},\vspace{0mm}
\end{equation}
where $\mathbf{M}_r$ is a relation-specific diagonal matrix, which represents the characteristics of a relation. Like triples in TransE, positive triples should have higher scores than negative ones in DistMult.


For DistMult and its extended models, we adopt the softplus loss function, which is shown as 
\begin{equation}\label{eq:softplus_loss} 
\mathcal{L}_{DistMult}=\sum_{(h^\prime,r,t^\prime)}\log(1+\exp(-y_{(h^\prime,r,t^\prime)}\cdot f(h^\prime,r,t^\prime))).
\end{equation}
In Equation (\ref{eq:softplus_loss}), $(h^\prime,r,t^\prime) \in \{(h,r,t)\}\cup Neg(h,r,t)$, where   $(h,r,t)$ is a selected triple in the training set, $Neg(h,r,t)\subset \left \{(h^\prime,r,t)|h^\prime \in \mathcal{E}\right \}\cup \left \{(h,r,t^\prime)|t^\prime \in \mathcal{E} \right \}$ is a set of corrupted triples. $y_{(h^\prime,r,t^\prime)}$ is the label of the triple $(h^\prime,r,t^\prime)$.

ConvE~\cite{dettmers2018convolutional} models the interactions between input entities and relations by convolutional and fully-connected layers. Given ($h$, $r$, $t$) triples, ConvE first reshapes the embedding of $h$ and $r$ into 2D tensors, then computes the scores of knowledge triples based on the reshaped tensors. The score function of ConvE is defined as
\begin{equation}\label{eq:conve_score}
f(h, r, t)=g\left ( \text{vec}(g( [ \hat{\mathbf{h}};\hat{\mathbf{r}}  ]\ast \mathbf{\omega} ))\mathbf{W} \right )\mathbf{t},\vspace{0mm}
\end{equation} 
where $\hat{\mathbf{h}}$ and $\hat{\mathbf{r}}$ are 2D reshapings of $\mathbf{h}$ and $\mathbf{r}$: if $\mathbf{h}, \mathbf{r} \in \mathbb{R}^d$, then $\hat{\mathbf{h}}, \hat{\mathbf{r}} \in \mathbb{R}^{d_1 \times  d_2}$, where $d = d_1d_2$. $\mathbf{\omega}$ denotes a set of filters and $\ast$ denotes the convolution operator.
$\text{vec}(\cdot)$ is a vectorization function, $g$ denotes the ReLU function, and $\mathbf{W}$ is the weight matrix.
ConvE also assumes that positive triples have higher scores than negative ones.

For ConvE and its extended models, we adopt the binary loss function, which is defined as 
\begin{align}
\mathcal{L}_{ConvE}=&\sum_{(h,r,t) \in \hat{\mathcal{T}}} -\frac{1}{N}\sum_{i=1}^N (y_{(h,r,t_i)}\cdot\log(\sigma(f(h,r,t_i)))\notag 
\\ & + (1-y_{(h,r,t_i)})\log(1- \sigma(f(h,r,t_i))) )\label{eq:binary_loss} ,
\end{align} 
where $\hat{\mathcal{T}}$ is the set of all the selected triples from the training set, $N$ denotes the number of  candidates for the tail entity, and $\sigma$ is the sigmoid function.

RotatE~\cite{sun2018rotate} is a recent model that maps entities and relations to the complex vector space and defines each relation as a rotation from the head entity to the tail entity. The score function of RotatE is defined as 
\begin{equation}\label{eq:rotate_score}
f(h, r, t) = -\left \| \mathbf{h} \circ \mathbf{r} - \mathbf{t} \right \|_{L_{1}},
\end{equation} 
where $\mathbf{h}, \mathbf{r}, \mathbf{t} \in \mathbb{C}^d$ are complex vectors, the modulus $|r_i| = 1$, and $\circ$ denotes the Hadamard (element-wise) product. Also, in RotatE, positive triples are supposed to have higher scores than negative ones.

The loss function of RotatE is defined as 
\begin{align}
\mathcal{L}_{RotatE}=&-\sum_{(h,r,t) \in \hat{\mathcal{T}}}[\log\sigma(f(h,r,t)-\eta) \notag 
\\ & - \Sigma_{(h^\prime,r,t^\prime)}\frac{1}{k}\log\sigma(\eta-f(h^\prime,r,t^\prime))]\label{eq:rotate_loss} ,
\end{align} 
where $\sigma(\cdot)$ is the sigmoid function, $k$ is the number of negative samples for each golden triple, and $\eta$ is a hyper-parameter.

\subsection{Model Training}
In our framework, before jointly training the policy-based agents and the KGE model, we pre-train the two modules respectively. First, we pre-train the KGE model based on the noisy training set.
It is worth noting that in order to prevent the KGE model overfitting on the noisy triples, the training epochs for the pre-training process of the KGE model is limited as 100.
Then the policy-based agents are pre-trained based on the embedding vectors from the KGE model. We find such a pre-training strategy is crucial for our framework, which is also widely used by previous reinforcement learning studies~\cite{qin2018robust,feng2018reinforcement}. After the pre-training process, the joint training procedure is conducted, which is described in Algorithm~\ref{algo:algo1}.
It is worth noting that we conducted experiments to explore the effect of triple order in the selection process on the final results. Experiments show that with the episode number $M$ (in Algorithm 1) getting larger, the effect of triple order is getting smaller. And in our setting, the order of triples has only a small effect on the final result (only up to 0.002 Mean Reciprocal Rank score). Thus we use a random order to conduct the experiments. 

\begin{algorithm}[!th]
	\SetAlgoNoLine
	\small
	\caption{Joint Training Procedure}
	\label{algo:algo1}
	\KwIn{Episode number $M$. A policy function and a KGE model parameterized by $\Theta$ and $\Phi$ respectively. A noisy training data set of the KGE model $\mathcal{T}=\{\mathcal{T}_r|r \in \mathcal{R}\}$\;}
	\KwOut{Entity embeddings \{$\mathbf{e}_1$, $\mathbf{e}_2$, ..., $\mathbf{e}_{|\mathcal{E}|}$\} and relation embeddings \{$\mathbf{r}_1$, $\mathbf{r}_2$, ..., $\mathbf{r}_{|\mathcal{R}|}$\} \;}
	\For{episode $m = 1$ to $M$}{
		\For {$r$ in $\mathcal{R}$}{
			Sample actions over $\mathcal{T}_r$ with $\Theta$: $A = \{a_1, ..., a_{|\mathcal{T}_r|}\}, a_t\sim\pi_\Theta(a_t|\mathbf{s}_t)$\;
			Get a cleansed triple set $\widehat{\mathcal{T}_r}$ for $r$ with $A$\;
			Replace the triples of relation $r$ in the training data set with $\widehat{\mathcal{T}_r}$\;
			Update $\Phi$ in the KGE model with the cleansed training data set \;
			Compute a reward $R$ with Equation (\ref{eq:reward})\;
			Update $\Theta$ with Equation (\ref{eq:gradients})\;
		}
	}
\end{algorithm} 
The entity and relation embeddings of the KGE models are initialized with a uniform distribution $U[ -6/\sqrt{d}, 6/\sqrt{d} ]$ following TransE~\cite{bordes2013translating}, where $d$ is the dimension of the embedding space. The learning process of the above models is carried out using the Adam optimizer~\cite{kingma2014adam}.
Specifically, for DistMult and its extended models, $L_2$ regularizer is applied to all the entity and relation embeddings during the training procedure.

%% file: experiments.tex
\section{Experiments}\label{sec:experiments}
In this section, we evaluate the proposed framework on the noise detection, the link prediction and the triple classification tasks.

\subsection{Dataset}\label{sec:datasets}

%

In the experiments, we use the datasets released by CKRL~\cite{xie2018does}. Specifically, in the paper of CKRL, three datasets are constructed with noisy triples to be 10\%, 20\% and 40\% of positive triples 
based on a popular benchmark FB15k~\cite{bordes2013translating}. All the three noisy datasets, which are denoted as FB15k-N1, FB15k-N2 and FB15k-N3 respectively, share the same entities, relations, validation and test sets with FB15k, with all generated negative triples fused into the original training set of FB15k.
Since most noise and conflicts in real-world KGs derive from the misunderstanding between similar entities, e.g., the noise (\textit{Donald Trump}, \textit{nationality}, \textit{Canada}) is more likely to occur in real-world KGs than (\textit{Donald Trump}, \textit{nationality}, \textit{Basketball}), the noise triples are generated in the following way. Given a positive entity ($h, r, t$), the head or tail entity is replaced to form a negative triple $(h^\prime, r, t)$ or $(h, r, t^\prime)$. The generation of negative triples is constrained that $h^\prime$ (or $t^\prime$) should have appeared in the head (or tail) position with the same relation $r$ in the dataset. Under this setting, the tail entity of the relation \textit{nationality} should be a country. This setting is able to generate harder and more confusing negative triples.

It is worth noting that recent studies have shown that FB15k suffers from the information leakage problem~\cite{toutanova2015observed,dettmers2018convolutional}. For example, the test set main contains triples (\textit{A}, \textit{contains}, \textit{B}) when the training set contains triple (\textit{B}, \textit{contained\_by}, \textit{A}). And one can attain the state-of-the-art results even using simple rules. In this case, we construct noisy datasets based on popular benchmark datasets FB15k-237~\cite{toutanova2015observed} and WN18RR~\cite{dettmers2018convolutional} using the same method introduced in CKRL~\cite{xie2018does}. The corresponding datasets are denoted as FB15k-237-N1, FB15k-237-N2 and FB15k-237-N3, WN18RR-N1, WN18RR-N2 and WN18RR-N3, respectively. The statistics of the datasets are summarized in Table~\ref{tab:fb15k_dataset}, Table \ref{tab:fb15k237_dataset} and Table \ref{tab:wn18rr_dataset}.

%

\begin{table}[t]
	\center 
	\small
	\caption{\label{tab:fb15k_dataset} Statistics of FB15k-based datasets.}
	\scalebox{1}{
		\begin{tabular}{cccccc}
			\toprule
			Dataset & \#Rel & \#Ent & \#Train & \#Valid & \#Test\\
			\midrule
			FB15k & 1,345 & 14,951 & 483,142 & 50,000 & 59,071\\
			\bottomrule
		\end{tabular}
	}
	\scalebox{1}{
		\begin{tabular}{p{48pt}<{\centering}p{42pt}<{\centering}p{42pt}<{\centering}p{42pt}<{\centering}}
			\toprule
			& FB15k-N1 & FB15k-N2 & FB15k-N3 \\
			\midrule
			\#Neg triple & 46,408 & 93,782 & 187,925 \\
			\bottomrule
		\end{tabular}
	}
\end{table}

\begin{table}[t]
	\center 
	\small
	\caption{\label{tab:fb15k237_dataset} Statistics of FB15k-237 based datasets.}
	\scalebox{.9}{
		\begin{tabular}{cccccc}
			\toprule
			Dataset & \#Rel & \#Ent & \#Train & \#Valid & \#Test\\
			\midrule
			FB15k-237 & 237 & 14,541 & 272,115 & 17,535 & 20,466\\
			\bottomrule
		\end{tabular}
	}
	\scalebox{.8}{
		\begin{tabular}{p{50pt}<{\centering}p{60pt}<{\centering}p{60pt}<{\centering}p{60pt}<{\centering}}
			\toprule
			& FB15k-237-N1 & FB15k-237-N2 & FB15k-237-N3 \\
			\midrule
			\#Neg triple & 27,211 & 54,423 & 108,846 \\
			\bottomrule
		\end{tabular}
	} 
\end{table}

\begin{table}[t]
	\center 
	\small
	\caption{\label{tab:wn18rr_dataset} Statistics of WN18RR based datasets.}
	\scalebox{.9}{
		\begin{tabular}{cccccc}
			\toprule
			Dataset & \#Rel & \#Ent & \#Train & \#Valid & \#Test\\
			\midrule
			WN18RR & 11 & 40,943 & 86,835 & 3,034 & 3,134\\
			\bottomrule
		\end{tabular}
	}
	\scalebox{.8}{
		\begin{tabular}{p{50pt}<{\centering}p{60pt}<{\centering}p{60pt}<{\centering}p{60pt}<{\centering}}
			\toprule
			& WN18RR-N1 & WN18RR-N2 & WN18RR-N3 \\
			\midrule
			\#Neg triple & 8,683 & 17,367 & 34,734 \\
			\bottomrule
		\end{tabular}
	}
\end{table}

\begin{figure*}[!tbp]
	\centering
	\subfigure[FB15k] {\label{fig:fb15k_noise_detection_bar}\includegraphics[width=4.5cm,height=6cm]{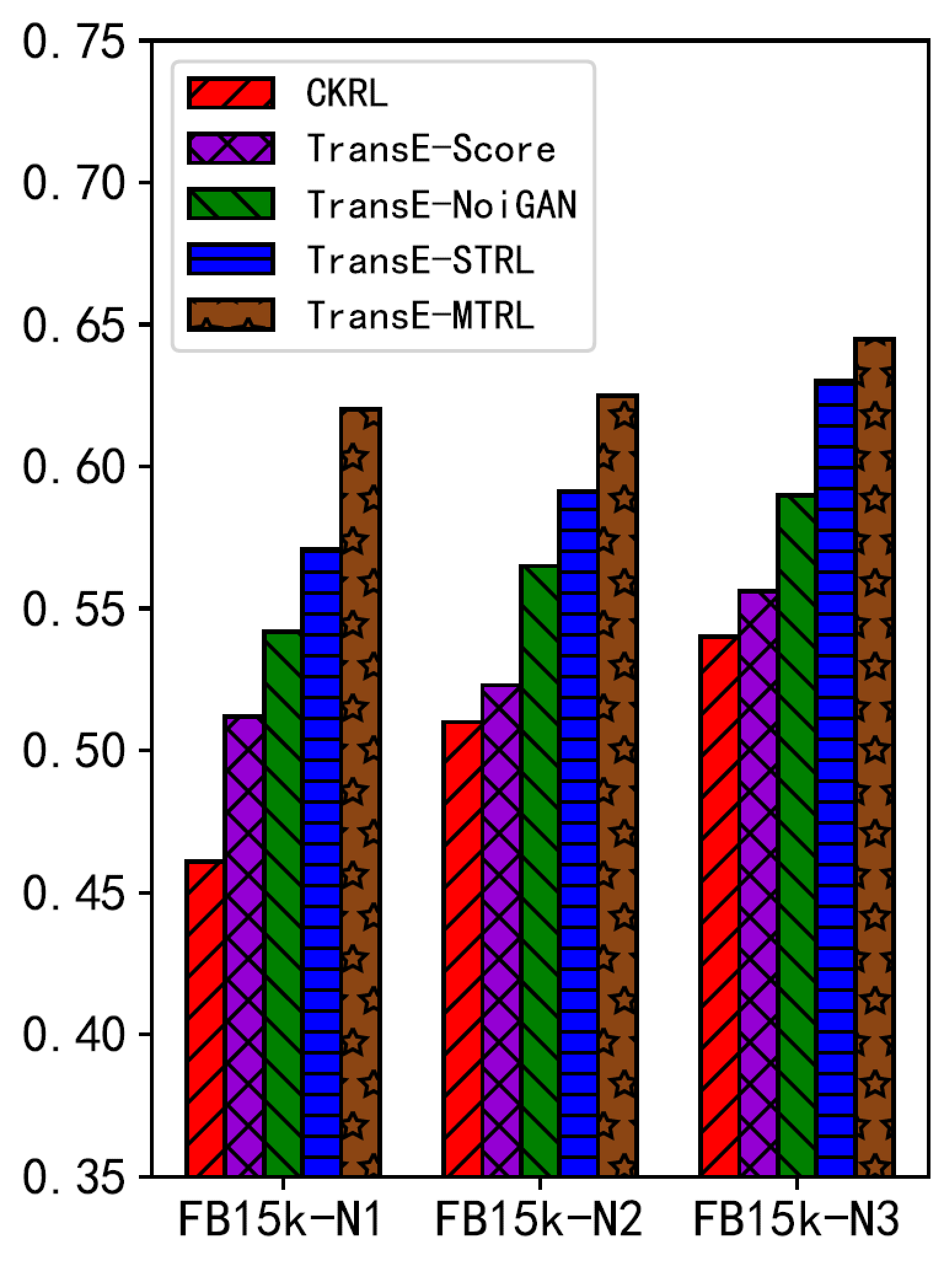}}	
	\subfigure[FB15k-237] {\label{fig:fb15k237_noise_detection_bar}\includegraphics[width=4.5cm,height=6cm]{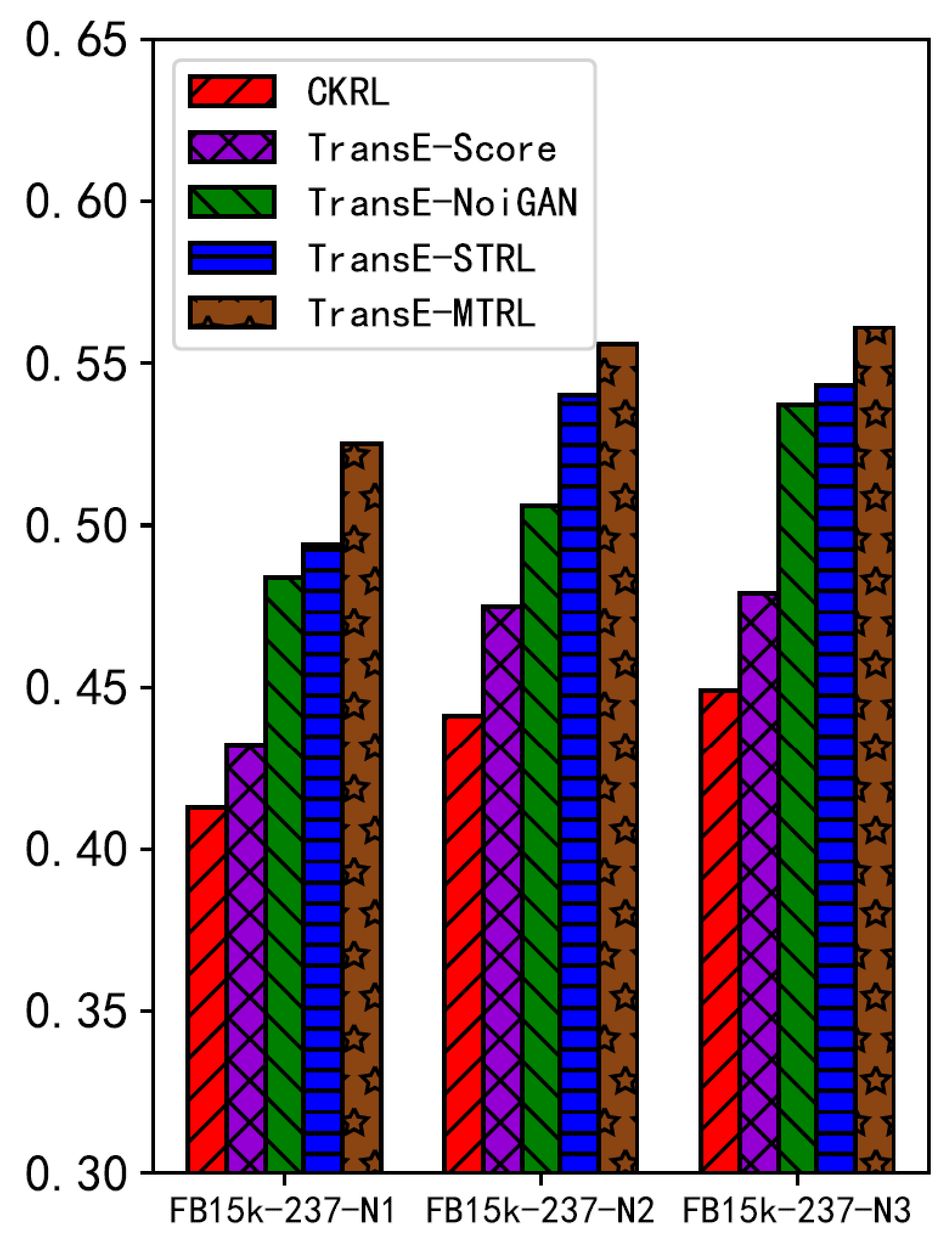}}
	\subfigure[WN18RR] {\label{fig:wn18rr_noise_detection_bar}\includegraphics[width=4.5cm,height=6cm]{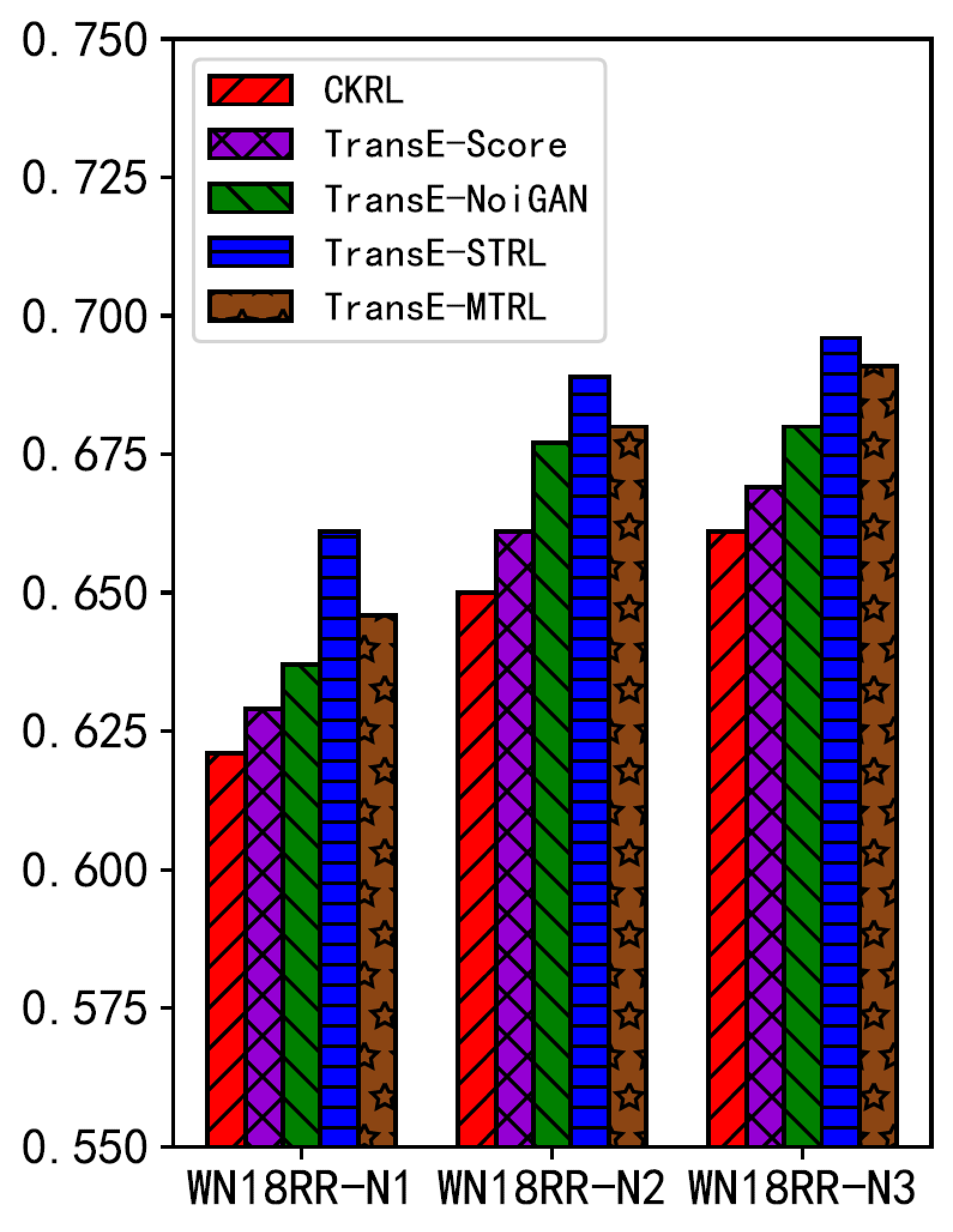}}
	\caption{KG noise detection results.}
	\label{fig:noise_detection}
\end{figure*}

\subsection{Baselines}

In this paper, we compare the proposed framework with the following baselines.
\begin{itemize}
	\item TransE~\cite{bordes2013translating}: one of the most popular and widely used translation-based models.
	\item DistMult~\cite{yang2015embedding}: a popular tensor factorization based model.
	\item ConvE~\cite{dettmers2018convolutional}: a state-of-the-art neural network based model, which utilizes convolutional neural networks to compute the scores of given triples.
	\item RotatE~\cite{sun2018rotate}: a recent KGE model which views each relation as a rotation from the head entity to the tail entity in the complex vector space.
	\item CKRL~\cite{xie2018does}: a state-of-the-art KGE model which computes a confidence score for each triple during the training procedure.
	\item X-Score: X-Score is a simple baseline that filters out triples with lower scores. Specifically, we first pre-train the KGE model with the noisy training set, then filter out the triples with lower scores. The proportion of triples that are removed is a hyperparameter $\delta$. Finally, we re-train the KGE model with the retained triples.
	\item X-NoiGAN: X-NoiGAN denotes the models extended with a recent confidence-aware method NoiGAN~\cite{anonymous2020noigan}. We use the hard version of NoiGAN since it achieves the best results. 
\end{itemize}

\subsection{Knowledge Graph Noise Detection}
The KG noise detection task aims to test the capacity of the reinforcement learning agents in selecting golden triples from the noisy training set. 

\subsubsection{Experimental Settings}

\begin{table}[th!]
	\centering
	\caption{Hyper-parameter Ranges.}\vspace{-2mm}
	\label{tab:parameter_range}
	\begin{tabular}{ll}
		\toprule
		Hyper-parameter & range  \\ \midrule
		$\alpha$ in Equation (\ref{eq:reward}) & \{0.02, 0.03, 0.04, 0.05\} \\ \midrule
		$\lambda_1$ and $\lambda_2$ in Equation (\ref{eq:omega}) & \{0.1, 0.01, 0.001\} \\ \midrule
		the number of episodes $M$  & \{5, 10, 15\}      \\ \midrule
		batch size  & \{256, 512, 1024\}  \\ \midrule
		learning rate  & \{0.1, 0.01, 0.001\}  \\ \midrule
		$\gamma$ in Equation (\ref{eq:margin_loss})  & \{1, 5, 10\} \\ \midrule
		$\eta$ in Equation (\ref{eq:rotate_loss})  & \{1, 5, 10\}  \\ \midrule
		$\delta$ in X-Score & \{0, 5\%, 10\%, ..., 1\} \\ \bottomrule
	\end{tabular}
\end{table}

\begin{table*}[t]
	\centering
	\caption{Link prediction results on FB15k-N1 to FB15k-N3.}\vspace{-2mm}
	\label{tab:kgc_results_fb15k}
	\scalebox{1.0}{
		\begin{tabular}{@{}lcccccccccccc@{}}
			\toprule
			& \multicolumn{4}{c}{FB15k-N1} & \multicolumn{4}{c}{FB15k-N2} & \multicolumn{4}{c}{FB15k-N3}
			\\
			\cmidrule(lr){2-5} \cmidrule(l){6-9} \cmidrule(l){10-13} & MRR & H@10 & H@3 & H@1 & MRR & H@10 & H@3 & H@1 & MRR & H@10 & H@3 & H@1  \\
			\midrule
			
			CKRL & 0.371 & 0.642 & 0.440 & 0.229 & 0.349 & 0.611 & 0.421 & 0.211 & 0.317 & 0.566 & 0.378 & 0.190 \\
			\midrule
			
			TransE & 0.365 & 0.636 & 0.442 & 0.222 & 0.346 & 0.605 & 0.415 & 0.210 & 0.314 & 0.557 & 0.372 & 0.188 \\
			TransE-Score & 0.368 & 0.638 & 0.432 & 0.224 & 0.347 & 0.607 & 0.417 & 0.209 & 0.317 & 0.557 & 0.378 & 0.187 \\
			TransE-NoiGAN & 0.372 & 0.639 & 0.443 & 0.231 & 0.349 & 0.612 & 0.424 & 0.214 & 0.322 & 0.568 & 0.387 & 0.213 \\
			TransE-STRL (Ours) & 0.373 & 0.642 & 0.445 & 0.226 & 0.359 & 0.621 & 0.429 & 0.218 & 0.328 & 0.577 & 0.391 & \textbf{0.215} \\
			TransE-MTRL (Ours) & \textbf{0.378*} & \textbf{0.644*} & \textbf{0.450*} & \textbf{0.235*} & \textbf{0.367*} & \textbf{0.628*} & \textbf{0.437*} & \textbf{0.228*} & \textbf{0.336*} & \textbf{0.582*} & \textbf{0.397*} & 0.214 \\
			\midrule
			
			DistMult & 0.435 & 0.700 & 0.517 & 0.298 & 0.361 & 0.653 & 0.428 & 0.222 & 0.251 & 0.528 & 0.281 & 0.128 \\
			DistMult-Score & 0.439 & 0.702 & 0.519 & 0.297 & 0.364 & 0.658 & 0.430 & 0.222 & 0.255 & 0.531 & 0.288 & 0.133 \\
			DistMult-NoiGAN & 0.441 & 0.708 & 0.527 & 0.299 & 0.366 & 0.661 & 0.435 & 0.226 & 0.259 & 0.537 & 0.295 & 0.141 \\
			DistMult-STRL (Ours) & 0.445 & \textbf{0.714} & 0.533 & 0.301 & 0.371 & 0.667 & 0.441 & 0.229 & 0.268 & 0.549 & 0.308 & 0.147 \\
			DistMult-MTRL (Ours) & \textbf{0.451*} & 0.713 & \textbf{0.539*} & \textbf{0.309*} & \textbf{0.379*} & \textbf{0.670*} & \textbf{0.443*} & \textbf{0.234*} & \textbf{0.283*} & \textbf{0.561*} & \textbf{0.313*} & \textbf{0.153*} \\
			\midrule
			
			ConvE & 0.513 & 0.711 & 0.578 & 0.402 & 0.499 & 0.699 & 0.562 & 0.387 & 0.470 & 0.672 & 0.533 & 0.359 \\
			ConvE-Score & 0.515 & 0.712 & 0.581 & 0.404 & 0.501 & 0.698 & 0.563 & 0.394 & 0.474 & 0.677 & 0.536 & 0.362 \\
			ConvE-NoiGAN & 0.518 & 0.714 & 0.579 & 0.407 & 0.503 & 0.702 & 0.565 & 0.396 & 0.483 & 0.684 & 0.543 & 0.372 \\
			ConvE-STRL (Ours) & 0.523 & 0.716 & 0.585 & 0.411 & 0.509 & 0.708 & 0.569 & 0.401 & 0.495 & 0.689 & 0.555 & 0.378 \\
			ConvE-MTRL (Ours) & \textbf{0.526*} & \textbf{0.723*} & \textbf{0.588*} & \textbf{0.415*} & \textbf{0.515*} & \textbf{0.715*} & \textbf{0.579*} & \textbf{0.405*} & \textbf{0.502*} & \textbf{0.699*} & \textbf{0.561*} & \textbf{0.389*} \\
			\midrule
			
			RotatE & 0.771 & 0.836 & 0.801 & 0.719 & 0.749 & 0.812 & 0.781 & 0.689 & 0.726 & 0.783 & 0.759 & 0.662 \\
			RotatE-Score & 0.772 & 0.838 & 0.804 & 0.726 & 0.751 & 0.819 & 0.784 & 0.689 & 0.733 & 0.789 & 0.764 & 0.669 \\
			RotatE-NoiGAN & 0.774 & 0.841 & 0.807 & 0.728 & 0.755 & 0.825 & 0.786 & 0.691 & 0.744 & 0.798 & 0.775 & 0.682 \\
			RotatE-STRL (Ours) & 0.779 & 0.841 & \underline{\textbf{0.816}} & 0.733 & 0.756 & 0.826 & 0.795 & 0.692 & 0.748 & 0.804 & \underline{\textbf{0.785}} & 0.679 \\
			RotatE-MTRL (Ours) & \underline{\textbf{0.783*}} & \underline{\textbf{0.849*}} & 0.813 & \underline{\textbf{0.736*}} & \underline{\textbf{0.764*}} & \underline{\textbf{0.831*}} & \underline{\textbf{0.798*}} & \underline{\textbf{0.706*}} & \underline{\textbf{0.755*}} & \underline{\textbf{0.815*}} & 0.781 & \underline{\textbf{0.689*}} \\
			\bottomrule
			
		\end{tabular}
	}
\end{table*}

\begin{table*}[t]
	\centering
	\caption{Link prediction results on FB15k-237-N1 to FB15k-237-N3.}\vspace{-2mm}
	\label{tab:kgc_results_fb15k237}
	\scalebox{1}{
		\begin{tabular}{@{}lcccccccccccc@{}}
			\toprule
			& \multicolumn{4}{c}{FB15k-237-N1} & \multicolumn{4}{c}{FB15k-237-N2} & \multicolumn{4}{c}{FB15k-237-N3}
			\\
			\cmidrule(lr){2-5} \cmidrule(l){6-9} \cmidrule(l){10-13} & MRR & H@10 & H@3 & H@1 & MRR & H@10 & H@3 & H@1 & MRR & H@10 & H@3 & H@1  \\
			\midrule
			
			CKRL & 0.227 & 0.387 & 0.249 & 0.144 & 0.209 & 0.371 & 0.234 & 0.133 & 0.195 & 0.359 & 0.222 & 0.129 \\
			\midrule
			
			TransE & 0.221 & 0.383 & 0.242 & 0.143 & 0.207 & 0.365 & 0.229 & 0.128 & 0.192 & 0.347 & 0.213 & 0.114 \\
			TransE-Score & 0.224 & 0.385 & 0.244 & 0.147 & 0.209 & 0.364 & 0.229 & 0.135 & 0.195 & 0.349  & 0.221 & 0.125 \\
			TransE-NoiGAN & 0.228 & 0.388 & 0.249 & 0.150 & 0.212 & 0.373 & 0.237 & 0.138 & 0.199 & 0.363 & 0.229 & 0.136 \\
			TransE-STRL (Ours) & 0.229 & 0.388 & 0.251 & \textbf{0.152} & 0.218 & 0.379 & 0.244 & 0.140 & 0.206 & 0.371 & 0.231 & 0.142 \\
			TransE-MTRL (Ours) & \textbf{0.231*} & \textbf{0.391*} & \textbf{0.255*} & 0.151 & \textbf{0.230*} & \textbf{0.388*} & \textbf{0.247*} & \textbf{0.147*} & \textbf{0.226*} & \textbf{0.382*} & \textbf{0.245*} & \textbf{0.144*} \\
			\midrule
			
			DistMult & 0.202 & 0.360 & 0.229 & 0.125 & 0.189 & 0.347 & 0.213 & 0.112 & 0.175 & 0.331 & 0.198 & 0.099 \\
			DistMult-Score & 0.202 & 0.365 & 0.226 & 0.126 & 0.191 & 0.349 & 0.219 & 0.117 & 0.181 & 0.334  & 0.206 & 0.102 \\
			DistMult-NoiGAN & 0.206 & 0.366 & 0.235 & 0.129 & 0.194 & 0.355 & 0.221 & 0.119 & 0.189 & 0.339 & 0.219 & 0.116 \\
			DistMult-STRL (Ours) & 0.209 & 0.371 & 0.234 & \textbf{0.133} & 0.199 & \textbf{0.365} & 0.223 & 0.120 & 0.192 & 0.351 & 0.223 & 0.119 \\
			DistMult-MTRL (Ours) & \textbf{0.214*} & \textbf{0.376*} & \textbf{0.239*} & 0.131 & \textbf{0.207*} & 0.363 & \textbf{0.233*} & \textbf{0.122*} & \textbf{0.201*} & \textbf{0.355*} & \textbf{0.230*} & \textbf{0.121*} \\
			\midrule
			
			ConvE & 0.242 & 0.391 & 0.261 & 0.159 & 0.229 & 0.381 & 0.247 & 0.142 & 0.212 & 0.364 & 0.228 & 0.131 \\
			ConvE-Score & 0.244 & 0.392 & 0.259 & 0.162 & 0.233 & 0.388 & 0.251 & 0.144 & 0.219 & 0.366  & 0.234 & 0.137 \\
			ConvE-NoiGAN & 0.249 & 0.399 & 0.267 & 0.168 & 0.238 & 0.392 & 0.259 & 0.149 & 0.229 & 0.375 & 0.258 & 0.147 \\
			ConvE-STRL (Ours) & 0.252 & 0.402 & 0.271 & 0.171 & 0.242 & \textbf{0.404} & 0.263 & 0.161 & 0.231 & 0.381 & 0.255 & 0.159\\
			ConvE-MTRL (Ours) & \textbf{0.258*} & \textbf{0.405*} & \textbf{0.274*} & \textbf{0.173*} & \textbf{0.252*} & 0.401 & \textbf{0.268*} & \textbf{0.166*} & \textbf{0.244*} & \textbf{0.392*} & \textbf{0.259*} & \textbf{0.162*} \\
			\midrule
			
			RotatE & 0.301 & 0.489 & 0.336 & 0.192 & 0.292 & 0.471 & 0.319 & 0.177 & 0.273 & 0.461 & 0.311 & 0.164 \\
			RotatE-Score & 0.303 & 0.488 & 0.335 & 0.197 & 0.295 & 0.477 & 0.324 & 0.179 & 0.278 & 0.464  & 0.319 & 0.169 \\
			RotatE-NoiGAN & 0.308 & 0.492 & 0.342 & 0.199 & 0.299 & 0.482 & 0.331 & 0.186 & 0.289 & 0.471 & 0.328 & 0.173 \\
			RotatE-STRL (Ours) & 0.311 & 0.497 & 0.342 & 0.202 & 0.299 & 0.489 & 0.329 & 0.191 & 0.289 & 0.471 & 0.326 & 0.185 \\
			RotatE-MTRL (Ours) & \underline{\textbf{0.318*}} & \underline{\textbf{0.501*}} & \underline{\textbf{0.349*}} & \underline{\textbf{0.205*}} & \underline{\textbf{0.311*}} & \underline{\textbf{0.492*}} & \underline{\textbf{0.338*}} & \underline{\textbf{0.201*}} & \underline{\textbf{0.302*}} & \underline{\textbf{0.491*}} & \underline{\textbf{0.331*}} & \underline{\textbf{0.199*}} \\
			\bottomrule
			
		\end{tabular}
	}
\end{table*}

\begin{table*}[t]
	\centering
	\caption{Link prediction results on WN18RR-N1 to WN18RR-N3.}\vspace{-2mm}
	\label{tab:kgc_results_wn18rr}
	\scalebox{1}{
		\begin{tabular}{@{}lcccccccccccc@{}}
			\toprule
			& \multicolumn{4}{c}{WN18RR-N1} & \multicolumn{4}{c}{WN18RR-N2} & \multicolumn{4}{c}{WN18RR-N3}
			\\
			\cmidrule(lr){2-5} \cmidrule(l){6-9} \cmidrule(l){10-13} & MRR & H@10 & H@3 & H@1 & MRR & H@10 & H@3 & H@1 & MRR & H@10 & H@3 & H@1  \\
			\midrule
			
			CKRL & 0.221 & 0.496 & 0.443 & 0.141 & 0.213 & 0.479 & 0.422 & 0.114 & 0.189 & 0.458 & 0.392 & 0.101 \\
			\midrule
			
			TransE & 0.219 & 0.489 & 0.431 & 0.137 & 0.201 & 0.471 & 0.414 & 0.112 & 0.179 & 0.446 & 0.389 & 0.094 \\
			TransE-Score & 0.221 & 0.493 & 0.435 & 0.141 & 0.207 & 0.473 & 0.419 & 0.114 & 0.183 & 0.445  & 0.392 & 0.098 \\
			TransE-NoiGAN & 0.225 & 0.499 & \textbf{0.447} & 0.143 & 0.219 & 0.477 & 0.425 & 0.117 & 0.194 & 0.461 & 0.399 & 0.106 \\
			TransE-STRL (Ours) & \textbf{0.231*} & \textbf{0.503*} & 0.446 & \textbf{0.146*} & \textbf{0.221*} & \textbf{0.486*} & \textbf{0.431*} & \textbf{0.124*} & \textbf{0.199*} & \textbf{0.473*} & \textbf{0.413*} & \textbf{0.122*} \\
			TransE-MTRL (Ours) & 0.226 & 0.499 & 0.445 & 0.145 & 0.219 & 0.479 & 0.428 & 0.116 & 0.196	& 0.468 & 0.399 & 0.113 \\
			\midrule
			
			DistMult & 0.411 & 0.473 & 0.425 & 0.376 & 0.394 & 0.455 & 0.413 & 0.352 & 0.371 & 0.437 & 0.392 & 0.329 \\
			DistMult-Score & 0.414 & 0.477 & 0.424 & 0.379 & 0.397 & 0.458 & 0.419 & 0.355 & 0.379 & 0.441 & 0.391 & 0.334 \\
			DistMult-NoiGAN & 0.419 & 0.481 & 0.428 & 0.379 & 0.405 & 0.463 & 0.424 & 0.366 & 0.392 & 0.455 & 0.412 & 0.336 \\
			DistMult-STRL (Ours) & \textbf{0.422*} & \textbf{0.485*} & \textbf{0.433*} & \textbf{0.385*} & \textbf{0.411*} & \textbf{0.472*} & \textbf{0.432*} & \textbf{0.371*} & \textbf{0.404*} & \textbf{0.461*} & \textbf{0.419*} & \textbf{0.353*} \\
			DistMult-MTRL (Ours) & 0.421 & 0.483 & 0.426 & 0.381 & 0.408 & 0.466 & 0.429 & 0.365 & 0.395 & 0.454 & 0.417 & 0.336 \\
			\midrule
			
			ConvE & 0.419 & 0.488 & 0.431 & 0.384 & 0.394 & 0.462 & 0.415 & 0.366 & 0.375 & 0.441 & 0.399 & 0.338 \\
			ConvE-Score & 0.421 & 0.491 & 0.433 & 0.383 & 0.396 & 0.465 & 0.414 & 0.369 & 0.383 & 0.451 & 0.402 & 0.341 \\
			ConvE-NoiGAN & 0.426 & 0.493 & \textbf{0.439} & 0.389 & 0.399 & 0.472 & 0.419 & 0.375 & 0.389 & 0.465 & 0.408 & 0.353 \\
			ConvE-STRL (Ours) & \textbf{0.431*} & \textbf{0.499*} & 0.436 & \textbf{0.396*} & \textbf{0.412*} & \textbf{0.479*} & \textbf{0.433*} & \textbf{0.383*} & \textbf{0.401*} & \textbf{0.473*} & \textbf{0.419*} & \textbf{0.361*} \\
			ConvE-MTRL (Ours) & 0.421 & 0.489 & 0.429 & 0.383 & 0.395 & 0.466 & 0.412 & 0.367 & 0.395 & 0.467 & 0.412 & 0.358 \\
			\midrule
			
			RotatE & 0.431 & 0.534 & 0.459 & 0.398 & 0.414 & 0.516 & 0.442 & 0.381 & 0.395 & 0.498 & 0.427 & 0.357 \\
			RotatE-Score & 0.431 & 0.533 & 0.461 & 0.403 & 0.417 & 0.519 & 0.448 & 0.385 & 0.401 & 0.503  & 0.431 & 0.364 \\
			RotatE-NoiGAN & 0.433 & 0.539 & 0.463 & \underline{\textbf{0.411}} & 0.429 & \underline{\textbf{0.538}} & 0.453 & 0.389 & 0.411 & 0.518 & 0.441 & 0.372 \\
			RotatE-STRL (Ours) & \underline{\textbf{0.439*}} & \underline{\textbf{0.547*}} & \underline{\textbf{0.471*}} & 0.409 & \underline{\textbf{0.433*}} & 0.537 & \underline{\textbf{0.461*}} & \underline{\textbf{0.396*}} & \underline{\textbf{0.426*}} & \underline{\textbf{0.522*}} & \underline{\textbf{0.452*}} & \underline{\textbf{0.385*}} \\
			RotatE-MTRL (Ours) & 0.433 & 0.541 & 0.467 & 0.405 & 0.432 & 0.536 & 0.458 & 0.393 & 0.417 & 0.518 & 0.447 & 0.381 \\
			\bottomrule
			
		\end{tabular}
	} 
\end{table*}

In the training stage, all the hyper-parameters are decided based on the model performance over the validation set via grid search. 
Specifically, in X-Score, a proportion of low-score triples are filtered out, and the proportion is denoted as a hyper-parameter $\delta$. Particularly, we use 5\% as the step size to set the value of $\delta$ from 0 to 1, and report the best performance. 
The hyper-parameter ranges are shown in Table~\ref{tab:parameter_range}. 
We find the best hyper-parameters as follows.
For the extended models of TransE, ConvE, and RotatE, $\alpha$ is set as 0.05, 0.03 and 0.02 for FB15k based and FB15k-237 based models, and 0.05, 0.04 and 0.02 for WN18RR based models. For the extended models of DistMult, $\alpha$ is set as 0.3, 0.2 and 0.1 for Y-N1 to Y-N3 respectively, where Y represents FB15k, FB15k-237 or WN18RR in this paper. For all the extended models, $\lambda_1$ and $\lambda_2$ are set as 0.001 and 0.01 respectively, and the episode number $M$ is set as 15. We set the batch size as 1024 for all the models. The learning rates are set as 0.001 for TransE, DistMult, ConvE and RotatE, and 0.0005 for their extended models. Specifically, for TransE and its extended models, $\gamma$ is set as 1 and the $L_1$ norm is adopted by the score function. For RotatE and its extended models, $\eta$ is set as 5. 
For X-Score, we set $\delta$ as 10\% on Y-N1, and 15\% on Y-N2. Specifically, for TransE-Score and RotatE-Score, $\delta$ is set as 25\% on Y-N3. And for DistMult-Score and ConvE-Score, $\delta$ is set as 30\% on Y-N3.
The number of relation clusters are set as 300, 120 and 10 for FB15k, FB15k-237 and WN18RR based datasets, respectively. 
On FB15k and FB15k-237 based datasets, as the number of relation clusters increases, the result first goes up and then falls down. On WN18RR based datasets, the model performance continues going up as the number of relation clusters increases.
For the impact of the number of relation clusters on the final results, please refer to the Appendix for details. We also tried different random seeds, but find the seed does not have a big effect on the final results (up to 0.001 MRR score).
The embedding size is set as 100 for all the models for a fair comparison.
For relations that have too many triples in the training data set (e.g., the relation \texttt{/people/person/profession} has 11972 triples in the training set of FB15k-237-N1), we sample a subset of 5000 triples to train the agents. 

In the test stage,  we adopt the F1 score to show the ability of our models in detecting noise. 
Specifically, for our models, since we make hard decisions over the noisy training set, the unselected triples are regarded as noise. For CKRL and X-Score, inspired by the evaluation method from the CKRL paper, the triples in the training set with lower $f(h,r,t)=-\left \| \mathbf{h}+\mathbf{r}-\mathbf{t} \right \|$ scores are detected as noisy ones, we compute the F1 scores with the recall changing from 0 to 1, and report the maximum F1 value. 

\subsubsection{Experimental Results}

The results are shown in Figure \ref{fig:noise_detection}. It can be clearly figured out that (1) on all the FB15k, FB15k-237, and WN18RR based datasets, TransE-STRL and TransE-MTRL clearly outperform CKRL, TransE-Score, and TransE-NoiGAN, which indicates that our framework can effectively detect and filter out noisy triples; (2) on FB15k based and FB15k-237 based data sets, TransE-MTRL outperforms TransE-STRL, which indicates that the information learned from similar relations is of great value, and validates the effectiveness of the collective training process for semantically similar relations. However, we find that on WN18RR based datasets, TransE-STRL achieves better results than TransE-MTRL. We conjecture the reason lies in that the semantic correlations among relations in FB15k and FB15k-237 are much stronger than that of WN18RR. Although the information learned from semantically similar relations are useful, the information learned from unrelated relations may damage the results~\cite{zhang2018knowledge}. The results are in line with the findings in \cite{zhang2018knowledge}, which shows that the MTRL model is more useful for KGs which have dense semantic distributions over relations, while the STRL model is more suitable for KGs in which the semantic correlations among relations are weak.

\subsection{Link Prediction}
Link prediction, a.k.a. knowledge graph completion, aims to fill the missing values into incomplete knowledge triples. More formally, the goal of link prediction is to predict either the head entity
in a given query ($?$, $r$, $t$) or the tail entity in a given query ($h$, $r$, $?$).

\subsubsection{Experimental Settings}
In the training phase, we use the same hyper-parameter settings as the KG noise detection task.
In the test phase, we replace the head and tail entities with all entities in KG in turn for each triple in the test set. Then we compute a score for each corrupted triple, and rank all the candidate entities according to the scores. Specifically, positive candidates are supposed to precede negative
ones. Finally, the rank of the correct entity is stored.
We compare our models with baselines using the following metrics: (i) Mean Reciprocal
Rank (MRR, the mean of all the reciprocals of predicted ranks);  (ii) Hits@$n$ (H@$n$, the proportion
of ranks not larger than $n$). All the results are reported in the ``filtered'' setting~\cite{bordes2013translating}.

\subsubsection{Experimental Results}

Evaluation results are shown in Table \ref{tab:kgc_results_fb15k}, Table \ref{tab:kgc_results_fb15k237} and Table~\ref{tab:kgc_results_wn18rr}. We divide all the results into 5 groups. The second, third, fourth and fifth group are the results of TransE, DistMult, ConvE, RotatE and their extended models, while the first group is the state-of-the-art baseline CKRL. 
Results in bold font are the best results in the group, and the underlined results denote the best results in the column. 
Numbers marked with * indicate that the improvement is statistically significant compared with the best baseline in the group (t-test with p-value \textless 0.05).
From these tables, we have the following findings. (1) Our extended models clearly outperform the base models and other competitors, which clearly validates the effectiveness of the proposed framework, and shows that the policy-based agents can well filter out the noisy triples and retain the positive ones.
(2) X-MTRL achieves the best results on FB15k and FB15k-237 based datasets, while X-STRL outperforms other baselines on WN18RR based data sets. The results confirm that MTRL models are more useful for KGs in which the semantic correlations among relations are strong, while the STRL models are more suitable for KGs which have sparse semantic distributions over relations.
(3) The margin between the base models and the best performed extended models become more significant as the noise rate in KGs goes higher. Taking the metric of MRR as an example, comparing with RotatE, RotatE-MTRL gets the improvements of 0.017, 0.019 and 0.029 on FB15k-237-N1 to FB15k-237-N3 respectively. It indicates the proposed framework can well handle KGs with different noise rates.

\begin{table*}[!t]
	\centering
	\caption{Triple selection examples by different models. The value indicates the weight of the triple in the training procedure.}\vspace{-2mm}
	\label{tab:case_study}
	\scalebox{1.0}{
		\begin{tabular}{@{}lccc@{}}
			\toprule
			Triple Instance & CKRL  & TransE-NoiGAN & TransE-MTRL\\
			\midrule
			(Brazil,  /location/location/contains,  Rio de Janeiro) & 0.84 & 1 & 1\\
			(Martin Luther King, /influence/influence\_node/influenced\_by, Abraham Lincoln) & 0.86 & 0 & 1 \\
			\midrule
			(Brazil, /location/location/contains, City of Toronto) & 0.14 & 0 & 0 \\
			(WA Mozart, /people/person/nationality, Japan) & 0.26 & 0 & 0 \\
			\bottomrule
		\end{tabular}
	}\vspace{-2mm}
\end{table*}

\begin{table*}[!t]
	\centering
	\caption{Examples of relation clusters in FB15k-237-N1.}\vspace{-2mm}
	\label{tab:clusters}
	\begin{tabular}{c|l}
		\hline
		& relations                                                                          \\ \hline
		\multirow{3}{*}{1}     & /film/film/produced\_by                                                            \\ \cline{2-2} 
		& /film/film/executive\_produced\_by                                                 \\ \cline{2-2} 
		& /film/film/film\_art\_direction\_by                                                \\ \hline
		\multirow{3}{*}{2}     & /soccer/football\_team/current\_roster./soccer/football\_roster\_position/position \\ \cline{2-2} 
		& /soccer/football\_team/current\_roster./sports/sports\_team\_roster/position       \\ \cline{2-2} 
		& /sports/sports\_position/players./sports/sports\_team\_roster/position             \\ \hline
		\multirow{2}{*}{3}     & /people/ethnicity/languages\_spoken                                                \\ \cline{2-2} 
		& /people/person/languages                                                           \\ \hline
	\end{tabular}\vspace{-2mm}
\end{table*}

We also provide some case studies.
Table~\ref{tab:case_study} shows some case studies of the triple selection process on FB15k-237-N1. In Table~\ref{tab:case_study}, four triple instances are divided into two groups. The first group contains two positive triples, and the second group is comprised of two negative ones. 
In Table~\ref{tab:case_study}, we normalize the confidence scores of CKRL to [0, 1] by $s_i = s_i / s_{max}$, where $s_i$ is the confidence score of the $i$-th triple and $s_{max}$ is the maximal confidence score. 
It clearly shows that our model TransE-MTRL can assign proper hard weights to triples. In addition, Table~\ref{tab:clusters} gives some examples of relation clusters in FB15k-237-N1. Relations in Cluster 1 are sports-related relations, in Cluster 2 are film-related relations, while in Cluster 3 are language-related relations. From Table~\ref{tab:clusters} we can see that semantically related relations are clustered into the same group, which is helpful to facilitate knowledge sharing among these relations.


To further evaluate the generalization performance of the proposed framework, we extend popular KGE models R-GCN~\cite{schlichtkrull2018modeling}, CompGCN~\cite{vashishth2020composition} and ComplEx-N3~\cite{lacroix2018canonical} with our framework. Due to the space limitation, please refer to the Appendix for more details. Experimental results validate the generalization ability of the proposed framework.

\subsubsection{Further Comparison between X-MTRL/STRL and X-Score}
To further evaluate the effectiveness of the reinforcement learning method in selecting golden triples, we compare X-MTRL/STRL and X-Score.
Specifically, we first run X-MTRL or X-STRL, and get the number of triples that are kept. Then, we run X-Score by keeping the same number of samples as X-MTRL or X-STRL. Finally, we compare the results of the above models. 
Particularly, since X-MTRL achieves the best results on FB15k and FB15k-237 based datasets, and X-STRL outperforms other models on WN18RR-based datasets, on FB15k and FB15k-237 based datasets, we compare X-Score with X-MTRL, while on WN18RR-based datasets, we compare X-Score with X-STRL.
The results are shown in Table~\ref{tab:comparison_results_fb15k}, Table~\ref{tab:comparison_results_fb15k237}, and Table~\ref{tab:comparison_results_wn18rr}.
We divide the results into 4 groups, which are the extended models of TransE, DistMult, ConvE, and RotatE, respectively.
Results in bold font are the best results in the group, and the underlined results denote the best results in the column. 
Numbers marked with * indicate that the improvement is statistically significant compared with the baseline in the group (t-test with p-value $<$ 0.05).
We observe that X-MTRL or X-STRL significantly outperforms X-Score with a large margin, which indicates that the reinforcement learning setting is capable of selecting better samples than the score-based method.

\begin{table*}[h]
	\centering
	\caption{Comparison between X-MTRL and X-Score on FB15k-N1 to FB15k-N3.}
	\label{tab:comparison_results_fb15k}
	\scalebox{1}{
		\begin{tabular}{@{}lcccccccccccc@{}}
			\toprule
			& \multicolumn{4}{c}{FB15k-N1} & \multicolumn{4}{c}{FB15k-N2} & \multicolumn{4}{c}{FB15k-N3}
			\\
			\cmidrule(lr){2-5} \cmidrule(l){6-9} \cmidrule(l){10-13} & MRR & H@10 & H@3 & H@1 & MRR & H@10 & H@3 & H@1 & MRR & H@10 & H@3 & H@1  \\
			\midrule
			
			TransE-Score & 0.362 & 0.631 & 0.424 & 0.221 & 0.341 & 0.598 & 0.414 & 0.203 & 0.313 & 0.548 & 0.372 & 0.179 \\
			TransE-MTRL (Ours) & \textbf{0.378*} & \textbf{0.644*} & \textbf{0.450*} & \textbf{0.235*} & \textbf{0.367*} & \textbf{0.628*} & \textbf{0.437*} & \textbf{0.228*} & \textbf{0.336*} & \textbf{0.582*} & \textbf{0.397*} & \textbf{0.214*} \\
			\midrule
			
			DistMult-Score & 0.433 & 0.695 & 0.515 & 0.293 & 0.355 & 0.651 & 0.422 & 0.221 & 0.251 & 0.527 & 0.285 & 0.128 \\
			DistMult-MTRL (Ours) & \textbf{0.451*} & \textbf{0.713*} & \textbf{0.539*} & \textbf{0.309*} & \textbf{0.379*} & \textbf{0.670*} & \textbf{0.443*} & \textbf{0.234*} & \textbf{0.283*} & \textbf{0.561*} & \textbf{0.313*} & \textbf{0.153*} \\
			\midrule
			
			ConvE-Score & 0.514 & 0.713 & 0.577 & 0.401 & 0.499 & 0.695 & 0.561 & 0.389 & 0.472 & 0.673 & 0.535 & 0.363 \\
			ConvE-MTRL (Ours) & \textbf{0.526*} & \textbf{0.723*} & \textbf{0.588*} & \textbf{0.415*} & \textbf{0.515*} & \textbf{0.715*} & \textbf{0.579*} & \textbf{0.405*} & \textbf{0.502*} & \textbf{0.699*} & \textbf{0.561*} & \textbf{0.389*} \\
			\midrule
			
			RotatE-Score & 0.768 & 0.834 & 0.802 & 0.723 & 0.744 & 0.818 & 0.785 & 0.686 & 0.731 & 0.788 & 0.758 & 0.664 \\
			RotatE-MTRL (Ours) & \underline{\textbf{0.783*}} & \underline{\textbf{0.849*}} & \underline{\textbf{0.813*}} & \underline{\textbf{0.736*}} & \underline{\textbf{0.764*}} & \underline{\textbf{0.831*}} & \underline{\textbf{0.798*}} & \underline{\textbf{0.706*}} & \underline{\textbf{0.755*}} & \underline{\textbf{0.815*}} & \underline{\textbf{0.781*}} & \underline{\textbf{0.689*}} \\
			\bottomrule
			
		\end{tabular}
	}
\end{table*}

\begin{table*}
	\centering
	\caption{Comparison between X-MTRL and X-Score on FB15k-237-N1 to FB15k-237-N3.}
	\label{tab:comparison_results_fb15k237}
	\scalebox{1}{
		\begin{tabular}{@{}lcccccccccccc@{}}
			\toprule
			& \multicolumn{4}{c}{FB15k-237-N1} & \multicolumn{4}{c}{FB15k-237-N2} & \multicolumn{4}{c}{FB15k-237-N3}
			\\
			\cmidrule(lr){2-5} \cmidrule(l){6-9} \cmidrule(l){10-13} & MRR & H@10 & H@3 & H@1 & MRR & H@10 & H@3 & H@1 & MRR & H@10 & H@3 & H@1  \\
			\midrule
			
			TransE-Score & 0.221 & 0.383 & 0.244 & 0.143 & 0.205 & 0.359 & 0.221 & 0.134 & 0.192 & 0.345  & 0.217 & 0.120 \\
			TransE-MTRL (Ours) & \textbf{0.231*} & \textbf{0.391*} & \textbf{0.255*} & \textbf{0.151*} & \textbf{0.230*} & \textbf{0.388*} & \textbf{0.247*} & \textbf{0.147*} & \textbf{0.226*} & \textbf{0.382*} & \textbf{0.245*} & \textbf{0.144*} \\
			\midrule
			
			DistMult-Score & 0.199 & 0.361 & 0.222 & 0.125 & 0.188 & 0.344 & 0.216 & 0.113 & 0.177 & 0.331  & 0.203 & 0.101 \\
			DistMult-MTRL (Ours) & \textbf{0.214*} & \textbf{0.376*} & \textbf{0.239*} & \textbf{0.131*} & \textbf{0.207*} & \textbf{0.363*} & \textbf{0.233*} & \textbf{0.122*} & \textbf{0.201*} & \textbf{0.355*} & \textbf{0.230*} & \textbf{0.121*} \\
			\midrule
			
			ConvE-Score & 0.241 & 0.391 & 0.255 & 0.158 & 0.230 & 0.384 & 0.244 & 0.141 & 0.218 & 0.364  & 0.231 & 0.133 \\
			ConvE-MTRL (Ours) & \textbf{0.258*} & \textbf{0.405*} & \textbf{0.274*} & \textbf{0.173*} & \textbf{0.252*} & \textbf{0.401*} & \textbf{0.268*} & \textbf{0.166*} & \textbf{0.244*} & \textbf{0.392*} & \textbf{0.259*} & \textbf{0.162*} \\
			\midrule
			
			RotatE-Score & 0.301 & 0.488 & 0.332 & 0.196 & 0.294 & 0.479 & 0.320 & 0.175 & 0.275 & 0.461  & 0.317 & 0.164 \\
			RotatE-MTRL (Ours) & \underline{\textbf{0.318*}} & \underline{\textbf{0.501*}} & \underline{\textbf{0.349*}} & \underline{\textbf{0.205*}} & \underline{\textbf{0.311*}} & \underline{\textbf{0.492*}} & \underline{\textbf{0.338*}} & \underline{\textbf{0.201*}} & \underline{\textbf{0.302*}} & \underline{\textbf{0.491*}} & \underline{\textbf{0.331*}} & \underline{\textbf{0.199*}} \\
			\bottomrule
			
		\end{tabular}
	}
\end{table*}

\begin{table*}[h]
	\centering
	\caption{Comparison between X-STRL and X-Score on WN18RR-N1 to WN18RR-N3.}
	\label{tab:comparison_results_wn18rr}
	\scalebox{1}{
		\begin{tabular}{@{}lcccccccccccc@{}}
			\toprule
			& \multicolumn{4}{c}{WN18RR-N1} & \multicolumn{4}{c}{WN18RR-N2} & \multicolumn{4}{c}{WN18RR-N3}
			\\
			\cmidrule(lr){2-5} \cmidrule(l){6-9} \cmidrule(l){10-13} & MRR & H@10 & H@3 & H@1 & MRR & H@10 & H@3 & H@1 & MRR & H@10 & H@3 & H@1  \\
			\midrule
			
			TransE-Score & 0.219 & 0.491 & 0.434 & 0.138 & 0.203 & 0.471 & 0.415 & 0.109 & 0.181 & 0.443  & 0.388 & 0.095 \\
			TransE-STRL (Ours) & \textbf{0.231*} & \textbf{0.503*} & \textbf{0.446*} & \textbf{0.146*} & \textbf{0.221*} & \textbf{0.486*} & \textbf{0.431*} & \textbf{0.124*} & \textbf{0.199*} & \textbf{0.473*} & \textbf{0.413*} & \textbf{0.122*} \\
			\midrule
			
			DistMult-Score & 0.411 & 0.475 & 0.423 & 0.375 & 0.395 & 0.453 & 0.417 & 0.351 & 0.375 & 0.437 & 0.388 & 0.331 \\
			DistMult-STRL (Ours) & \textbf{0.422*} & \textbf{0.485*} & \textbf{0.433*} & \textbf{0.385*} & \textbf{0.411*} & \textbf{0.472*} & \textbf{0.432*} & \textbf{0.371*} & \textbf{0.404*} & \textbf{0.461*} & \textbf{0.419*} & \textbf{0.353*} \\
			\midrule
			
			ConvE-Score & 0.418 & 0.488 & 0.430 & 0.381 & 0.397 & 0.461 & 0.409 & 0.365 & 0.383 & 0.450 & 0.402 & 0.339 \\
			ConvE-STRL (Ours) & \textbf{0.431*} & \textbf{0.499*} & \textbf{0.436*} & \textbf{0.396*} & \textbf{0.412*} & \textbf{0.479*} & \textbf{0.433*} & \textbf{0.383*} & \textbf{0.401*} & \textbf{0.473*} & \textbf{0.419*} & \textbf{0.361*} \\
			\midrule
			
			RotatE-Score & 0.429 & 0.532 & 0.458 & 0.401 & 0.418 & 0.517 & 0.445 & 0.381 & 0.399 & 0.505  & 0.428 & 0.361 \\
			RotatE-STRL (Ours) & \underline{\textbf{0.439*}} & \underline{\textbf{0.547*}} & \underline{\textbf{0.471*}} & \underline{\textbf{0.409*}} & \underline{\textbf{0.433*}} & \underline{\textbf{0.537*}} & \underline{\textbf{0.461*}} & \underline{\textbf{0.396*}} & \underline{\textbf{0.426*}} & \underline{\textbf{0.522*}} & \underline{\textbf{0.452*}} & \underline{\textbf{0.385*}} \\
			\bottomrule
			
		\end{tabular}
	} 
\end{table*}

\subsubsection{Link Prediction Results on Sparse KGs}
To evaluate the model performance on sparse KGs, we manually remove triples from the original training set of FB15k-237-N1, and construct four sparser datasets, which are shown in Table~\ref{tab:sparse_kg_statistics}. The construction process is as follows, we first sample 80\%, 60\%, 40\% and 20\% of positive triples from the original training set of FB15k-237, then add noise to the sampled datasets with the same noise injection method as FB15k-237-N1. The four datasets are named as FB15k-237-N1-80\%, FB15k-237-N1-60\%, FB15k-237-N1-40\% and FB15k-237-N1-20\% respectively with different degrees of sparsity. 

\begin{table}[th!]
	\centering
	\caption{Statistics of Datasets.}
	\label{tab:sparse_kg_statistics}
	\begin{tabular}{ll}
		\toprule
		Datasets & \#training triples  \\ \midrule
		FB15k-237-N1 & 299,326 \\ \midrule
		FB15k-237-N1-80\% & 239,460 \\ \midrule
		FB15k-237-N1-60\%  & 179,595      \\ \midrule
		FB15k-237-N1-40\%  & 119,730  \\ \midrule
		FB15k-237-N1-20\%  & 59,865  \\ \bottomrule
	\end{tabular}
\end{table}

The results are shown in Table~\ref{tab:kgc_results_sparse}. Results in bold font are the best results in the group, and the underlined results denote the best results in the column. 
Numbers marked with * indicate that the improvement is statistically significant compared with the best baseline in the group (t-test with p-value \textless 0.05).
From Table~\ref{tab:kgc_results_sparse}, we have the following findings. (1) Our extended models outperform the base models and other baselines. And MTRL models achieve better results than STRL ones. The results confirm the effectiveness of our MTRL framework. (2) We find our extended models outperform the base models on KGs with different degrees of sparsity, which clearly validates that the proposed framework is able to achieve better results on both dense and sparse KGs. 

\begin{table*}
	\centering
	\caption{Link prediction results on sparse KGs.}
	\label{tab:kgc_results_sparse}
	\centering	
	\scalebox{.7}{
		\begin{tabular}{@{}lcccccccccccccccccccc@{}}
			\toprule
			& \multicolumn{4}{c}{FB15k-237-N1} & \multicolumn{4}{c}{FB15k-237-N1-80\%} & \multicolumn{4}{c}{FB15k-237-N1-60\%} & \multicolumn{4}{c}{FB15k-237-N1-40\%} & \multicolumn{4}{c}{FB15k-237-N1-20\%}
			\\
			\cmidrule(lr){2-5} \cmidrule(l){6-9} \cmidrule(l){10-13} \cmidrule(l){14-17} \cmidrule(l){18-21} & MRR & H@10 & H@3 & H@1 & MRR & H@10 & H@3 & H@1 & MRR & H@10 & H@3 & H@1 & MRR & H@10 & H@3 & H@1 & MRR & H@10 & H@3 & H@1 \\
			\midrule
			
			CKRL & 0.227 & 0.387 & 0.249 & 0.144 & 0.193 & 0.324 & 0.218 & 0.126 & 0.144 & 0.246 & 0.152 & 0.107 & 0.115 & 0.179 & 0.103 & 0.078 & 0.047 & 0.083 & 0.052 & 0.026  \\
			\midrule
			
			TransE & 0.221 & 0.383 & 0.242 & 0.143 & 0.191 & 0.322 & 0.214 & 0.124 & 0.147 & 0.251 & 0.155 & 0.115 & 0.122 & 0.181 & 0.105 & 0.089 & 0.059 & 0.089 & 0.056 & 0.031 \\
			TransE-NoiGAN & 0.228 & 0.388 & 0.249 & 0.150 & 0.199 & 0.329 & 0.221 & 0.131 & 0.149 & 0.262 & 0.162 & \textbf{0.119} & 0.126 & 0.183 & 0.105 & 0.086 & 0.063 & 0.088 & 0.059 & 0.033 \\
			TransE-STRL (Ours) & 0.229 & 0.388 & 0.251 & \textbf{0.152} & 0.207 & 0.335 & 0.224 & 0.136 & 0.154 & 0.271 & 0.168 & 0.114 & 0.132 & 0.188 & 0.109 & 0.086 & 0.066 & 0.087 & 0.064 & 0.032 \\
			TransE-MTRL (Ours) & \textbf{0.231*} & \textbf{0.391*} & \textbf{0.255*} & 0.151 & \textbf{0.211*} & \textbf{0.342*} & \textbf{0.231*} & \textbf{0.144*} & \textbf{0.158*} & \textbf{0.277*} & \textbf{0.181*} & 0.117 & \textbf{0.137*} & \textbf{0.191*} & \textbf{0.114*} & \textbf{0.098*} & \textbf{0.073*} & \textbf{0.097*} & \textbf{0.071*} & \textbf{0.039*} \\
			\midrule
			
			DistMult & 0.202 & 0.360 & 0.229 & 0.125 & 0.173 & 0.305 & 0.196 & 0.101 & 0.143 & 0.256 & 0.162 & 0.089 & 0.103 & 0.163 & 0.089 & 0.067 & 0.033 & 0.075 & 0.047 & 0.009 \\
			DistMult-NoiGAN & 0.206 & 0.366 & 0.235 & 0.129 & 0.173 & 0.311 & 0.199 & 0.106 & 0.146 & 0.251 & 0.166 & 0.091 & 0.109 & 0.166 & 0.091 & 0.069 & \textbf{0.036} & \textbf{0.087} & 0.049 & 0.014 \\
			DistMult-STRL (Ours) & 0.209 & 0.371 & 0.234 & \textbf{0.133} & 0.177 & 0.311 & 0.206 & 0.104 & 0.151 & 0.259 & 0.167 & 0.099 & 0.112 & \textbf{0.172} & 0.092 & 0.072 & 0.034 & 0.079 & 0.045 & 0.018 \\
			DistMult-MTRL (Ours) & \textbf{0.214*} & \textbf{0.376*} & \textbf{0.239*} & 0.131 & \textbf{0.185*} & \textbf{0.319*} & \textbf{0.215*} & \textbf{0.113*} & \textbf{0.154*} & \textbf{0.263*} & \textbf{0.171*} & \textbf{0.104*} & \textbf{0.118*} & 0.169 & \textbf{0.099*} & \textbf{0.076*} & 0.034 & 0.083 & \textbf{0.053*} & \textbf{0.022*} \\
			\midrule
			
			ConvE & 0.242 & 0.391 & 0.261 & 0.159 & 0.219 & 0.341 & 0.219 & 0.142 & 0.181 & 0.279 & 0.198 & 0.128 & 0.143 & 0.224 & 0.147 & 0.098 & 0.077 & 0.103 & 0.072 & 0.056 \\
			ConvE-NoiGAN & 0.249 & 0.399 & 0.267 & 0.168 & 0.222 & 0.344 & 0.229 & 0.144 & 0.183 & 0.283 & 0.201 & 0.131 & 0.144 & 0.231 & 0.154 & 0.102 & 0.078 & 0.104 & 0.074 & 0.059 \\
			ConvE-STRL (Ours) & 0.252 & 0.402 & 0.271 & 0.171 & 0.229 & 0.349 & \textbf{0.238} & 0.154 & 0.189 & 0.288 & 0.203 & 0.133 & 0.154 & 0.239 & 0.161 & 0.105 & 0.082 & \textbf{0.112} & 0.079 & 0.063 \\
			ConvE-MTRL (Ours) & \textbf{0.258*} & \textbf{0.405*} & \textbf{0.274*} & \textbf{0.173*} & \textbf{0.231*} & \textbf{0.351*} & 0.231 & \textbf{0.162*} & \textbf{0.192*} & \textbf{0.292*} & \textbf{0.209*} & \textbf{0.139*} & \textbf{0.161*} & \textbf{0.245*} & \textbf{0.169*} & \textbf{0.111*} & \textbf{0.084*} & 0.109 & \textbf{0.083*} & \textbf{0.068*} \\
			\midrule
			
			RotatE & 0.301 & 0.489 & 0.336 & 0.192 & 0.251 & 0.402 & 0.279 & 0.171 & 0.225 & 0.335 & 0.242 & 0.154 & 0.174 & 0.243 & 0.163 & 0.114 & 0.087 & 0.119 & 0.083 & 0.071 \\
			RotatE-NoiGAN & 0.308 & 0.492 & 0.342 & 0.199 & 0.258 & 0.409 & 0.283 & 0.174 & 0.228 & \underline{\textbf{0.343}} & 0.245 & 0.155 & 0.176 & 0.246 & 0.169 & 0.121 & 0.088 & \underline{\textbf{0.124}} & 0.091 & 0.067 \\
			RotatE-STRL (Ours) & 0.311 & 0.497 & 0.342 & 0.202 & 0.264 & 0.405 & 0.291 & \underline{\textbf{0.179}} & 0.231 & 0.339 & 0.248 & 0.157 & 0.176 & 0.249 & 0.168 & \underline{\textbf{0.132}} & 0.088 & 0.123 & 0.092 & 0.066 \\
			RotatE-MTRL (Ours) & \underline{\textbf{0.318*}} & \underline{\textbf{0.501*}} & \underline{\textbf{0.349*}} & \underline{\textbf{0.205*}} & \underline{\textbf{0.269*}} & \underline{\textbf{0.417*}} & \underline{\textbf{0.294*}} & 0.177 & \underline{\textbf{0.233*}} & 0.342 & \underline{\textbf{0.253*}} & \underline{\textbf{0.161*}} & \underline{\textbf{0.185*}} & \underline{\textbf{0.255*}} & \underline{\textbf{0.177*}} & 0.128 & \underline{\textbf{0.094*}} & 0.123 & \underline{\textbf{0.102*}} & \underline{\textbf{0.074*}} \\
			\bottomrule
			
		\end{tabular}
	}
\end{table*}

\begin{table*}[t]
	\centering
	\caption{Triple Classification Results.}\vspace{-2mm}
	\label{tab:triple_classification_results}
	\scalebox{1.0}{
		\begin{tabular}{@{}lccc@{}}
			\toprule
			Model & FB15k-N1/N2/N3 & FB15k-237-N1/N2/N3 & WN18RR-N1/N2/N3\\
			\midrule
			
			CKRL & 75.6 / 74.3 / 72.6 & 81.7 / 80.2 / 78.3 & 82.5 / 81.3 / 79.4 \\
			\midrule
			TransE & 75.1 / 73.9 / 72.2 & 81.5 / 79.6 / 77.3 & 81.9 / 80.7 / 78.8 \\
			TransE-Score & 75.2 / 73.9 / 72.6 & 81.7 / 79.9 / 77.9 & 82.1 / 81.2 / 79.8 \\
			TransE-NoiGAN & 75.6 / 75.1 / 73.2 & 81.7 / 80.4 / 78.9 & 82.4 / 81.4 / 79.9 \\
			TransE-STRL (Ours) & 75.9 / 75.4 / 74.2 & 81.9 / 80.5 / 78.9 & \textbf{82.9*} / \textbf{82.5*} / \textbf{81.1*} \\
			TransE-MTRL (Ours) & \textbf{76.3*} / \textbf{75.6*} / \textbf{74.6*} & \textbf{82.1*} / \textbf{81.4*} / \textbf{79.8*} & 82.5 / 81.3 / 80.4 \\
			\midrule
			
			DistMult & 74.2 / 73.5 / 71.9 & 84.1 / 83.3 / 80.6 & 82.5 / 81.4 / 79.6 \\
			DistMult-Score & 74.4 / 73.6 / 72.0 & 84.2 / 83.5 / 81.4 & 82.5 / 81.7 / 80.2 \\
			DistMult-NoiGAN & 74.4 / 74.4 / 72.3 & 84.8 / 84.4 / 81.8 & \textbf{83.6} / 82.9 / 81.9 \\
			DistMult-STRL (Ours) & 74.8 / 74.6 / 73.4 & 84.7 / 84.1 / 82.2 & 83.4 / \textbf{83.1*} / \textbf{82.7*} \\
			DistMult-MTRL (Ours) & \textbf{74.9*} / \textbf{74.7*} / \textbf{74.1*} & \textbf{85.5*} / \textbf{85.2*} / \textbf{82.7*} & 82.9 / 82.8 / 82.1 \\
			\midrule
			ConvE & 77.8 / 76.9 / 75.4 & 86.9 / 85.5 / 81.3 & 82.9 / 82.1 / 80.8 \\
			ConvE-Score & 77.9 / 77.2 / 75.9 & 87.2 / 85.9 / 82.1 & 82.8 / 82.3 / 81.4 \\
			ConvE-NoiGAN & 77.8 / 77.6 / 76.8 & 87.2 / 86.2 / 83.3 & 82.9 / 82.8 / 81.9 \\
			ConvE-STRL (Ours) & 78.3 / 77.8 / 77.1 & 87.4 / 86.5 / 83.6 & \textbf{83.8*} / \textbf{83.6*} / \textbf{83.1*} \\
			ConvE-MTRL (Ours) & \textbf{78.5*} / \textbf{78.1*} / \textbf{77.4*} & \textbf{87.5*} / \textbf{86.8*} / \textbf{84.8*} & 83.4 / 83.1 / 82.4 \\
			\midrule
			RotatE & 78.8 / 77.5 / 76.1 & 87.5 / 86.1 / 82.9 & 84.1 / 83.6 / 81.9 \\
			RotatE-Score & 79.0 / 77.8 / 76.6 & 87.8 / 86.2 / 83.5 & 84.3 / 83.9 / 82.5 \\
			RotatE-NoiGAN & 79.6 / 77.9 / 77.4 & 87.9 / 86.8 / 83.9 & 84.3 / 83.9 / 82.7 \\
			RotatE-STRL (Ours) & 79.6 / 78.8 / 77.9 & 87.9 / 87.1 / 84.1 & \underline{\textbf{84.7*}} / \underline{\textbf{84.5*}} / \underline{\textbf{83.9*}} \\
			RotatE-MTRL (Ours) & \underline{\textbf{79.9*}} / \underline{\textbf{79.1*}} / \underline{\textbf{78.5*}} & \underline{\textbf{88.3*}} / \underline{\textbf{87.9*}} / \underline{\textbf{85.3*}} & 84.4 / 84.1 / 83.1 \\
			\bottomrule
			
		\end{tabular}
	}\vspace{-2mm}
\end{table*}

\subsection{Triple Classification}
The triple classification task aims to predict the label (True or False) of a given triple ($h$, $r$, $t$), which is a benchmark test that evaluates the discriminative capability of KGE models.

\subsubsection{Experimental Settings}

All the hyper-parameters are set the same as those introduced in the KG noise detection task. Since there are no explicit negative triples in existing KGs, we construct negative triples in validation and test set following the same protocol as described in Section~\ref{sec:datasets}. In the test phase, we follow the same decision
process as introduced in~\cite{socher2013reasoning}: for TransE, DistMult, ConvE, RotatE and their extended models, a triple is regarded as a positive one if $f$($h$, $r$, $t$) is above a threshold; otherwise negative. The thresholds are determined on the validation set. We adopt accuracy as our evaluation metric.

\subsubsection{Experimental Results}
The experimental results are shown in Table \ref{tab:triple_classification_results}, which are also divided into 5 groups in the same way as Table \ref{tab:kgc_results_fb15k}, Table \ref{tab:kgc_results_fb15k237} and Table~\ref{tab:kgc_results_wn18rr}. 
Numbers marked with * indicate that the improvement is statistically significant compared with the best baseline in the group (t-test with p-value \textless 0.05).
Table \ref{tab:triple_classification_results} leads to the following conclusions: 
(1) Our extended models achieve better results than baselines, which shows the discriminative capability of the extended models, and again validates the effectiveness and extendibility of our framework. 
(2) We also find that MTRL models achieve the best results on FB15k and FB15k-237 based datasets, and STRL models perform the best in WN18RR based datasets, which again validates the MTRL models and SRTL models are more suitable for KGs with dense and sparse semantic distributions over relations, respectively.
(3) The results confirm the quality of the learned knowledge representations, since they not only outperform baselines in the link prediction task, but also achieve better results in the triple classification task.

%% file: conclusion.tex
\section{Conclusion}\label{sec:conclusion}
In this paper, we proposed a general multi-task reinforcement learning framework for robust KGE. Specifically,  we exploited  reinforcement learning to select positive triples for each relation and utilized multi-task learning to facilitate knowledge sharing among semantically similar relations. Moreover, we extended four
popular KGE models with the proposed framework. Finally, we evaluated our framework on the KG noise detection, the link prediction and the triple classification tasks. The results showed that our approach could enhance existing KGE models and provide more robust representations of KGs. We are hopeful that our framework can provide a new perspective for other noise-aware tasks.